# Can saliency maps flag chest radiograph classifier errors at a new hospital? An external validation study

Aashish Chapagai[1*], Angad Singh Gurtatta[1], Erik Cambria[1]

[1]College of Computing and Data Science, Nanyang Technological University, Singapore, Singapore.
[*]Corresponding author. E-mail: contact@aashishchapagai.com

**Abstract**

Saliency maps are offered as a safeguard for medical image classifiers, but whether they reveal individual errors at a new hospital remains largely untested. We trained a convolutional network and a vision transformer on NIH ChestX-ray14 and evaluated 303,768 decisions per model at the training hospital and at two external sites, in the USA and Vietnam, with radiologist-consensus labels. We scored each explanation by its anatomical enrichment, the area-normalized concentration of attribution in a segmented thoracic region, and tested whether it detects errors beyond the model's confidence. Grad-CAM localized findings less often than radiologists (hit rate at most 0.575 versus 0.888). Adding enrichment to a confidence-based error detector calibrated at the training hospital never raised its area under the curve by more than 0.011, and for the convolutional network the relation between enrichment and correctness reversed between hospitals, degrading the transferred detector. Model confidence, not saliency, should anchor deployment monitoring.

## Introduction

Deep learning models trained on large hospital datasets such as NIH ChestX-ray14[1] read chest radiographs about as well as radiologists when tested on held-out data from the same hospitals[2,3]. Moved to another hospital, they often do worse, partly because they learn acquisition-specific and site-specific shortcuts instead of disease[4–7]. Many of the resulting errors are silent: a wrong output looks no less confident than a correct one[8].

Saliency maps are the usual answer to this opacity. A heatmap is shown beside each prediction on the assumption that a reader will notice when the model looks in the wrong place, and international consensus guidance for trustworthy AI in health care lists explainability among its six guiding principles[9]. The evidence that heatmaps can carry this weight is thin. They localize findings less accurately than radiologists[10], several fail tests of localization, robustness and repeatability[11,12], some barely change when a model's weights are randomized[13], and a plausible map need not reveal why a model generalizes poorly[14]. Explanations can also mislead. In a randomized study of clinicians, showing explanations did not offset the harm done by a systematically biased model[15], and post hoc explanations have been argued to offer false reassurance about individual patients[16].

These studies ask whether a map looks right. A deployed system needs something narrower: a signal, available without labels, that a particular prediction is probably wrong. This is the failure-detection problem[17–19]. In medical imaging it is stubborn, because no confidence score consistently outperforms simple baselines across tasks and shifts[20,21], and whether explanations add anything, particularly after a move to a new hospital, is unknown. Related work has summarized attribution inside the lungs at cohort level under domain shift[22], tracked saliency localization from NIH ChestX-ray14 to VinDr-CXR[23], audited whether saliency overlaps annotated findings within one dataset[24], combined explanation quality with confidence to decide when to defer within one dataset[25], and detected distribution shift or out-of-distribution inputs from explanations[26,27]. To our knowledge, none has tested whether explanations detect individual errors better than confidence does under real cross-hospital shift.

We ran that test at scale (Fig. 1). We trained a convolutional network (DenseNet-121[28]) and a vision transformer (ViT-B/16[29]) on NIH ChestX-ray14 and evaluated every decision on 25,554 test radiographs from three sites in two countries: the NIH internal test set; Stanford Hospital, using CheXpert radiographs with radiologist segmentations and a radiologist localization benchmark[3,10]; and two hospitals in Hanoi, using VinDr-CXR radiographs with five-radiologist consensus labels and boxes[30]. We summarized Grad-CAM, LIME and SHAP explanations by their anatomical enrichment, the area-normalized concentration of attribution in a thoracic region segmented from each radiograph, and prespecified three hypotheses: that explanations of the same finding become less concentrated on the chest at a new hospital (H1), that false-positive findings have less concentrated explanations than true positives (H2), and that adding enrichment to the model's confidence improves failure detection at new hospitals (H3). We find that explanations shift between hospitals but do not identify errors, that the relation between explanation plausibility and correctness depends on the hospital and, for the convolutional network, reverses between hospitals, and that for networks with a global-average-pooling head the class activation map is not independent evidence about the decision it explains, because it is an exact rearrangement of the class score.

## Results

### Study design, cohorts and flags

The two classifiers reached internal macro areas under the receiver operating characteristic curve (AUROCs) of 0.847 (DenseNet-121; 95% confidence interval (CI) 0.838–0.854) and 0.837 (ViT-B/16; 0.830–0.844) across 14 findings (Extended Data Table 1). We harmonized 12 findings across the three label sets (Table 1). Each radiograph–finding pair is one decision, giving 303,768 decisions per model, and a finding is flagged when the margin between its logit and the logit of its NIH-tuned threshold is positive (Methods). Flags were often wrong: flag precision was 0.358 (DenseNet-121) and 0.343 (ViT-B/16) at NIH, 0.529 and 0.514 at CheXpert, and 0.434 and 0.370 at VinDr-CXR, so between 47% and 66% of flags were false positives. Detecting these errors without labels is the task we study. To fit score combinations without leakage, NIH test patients were split at random into a calibration half (3,080 patients) and an evaluation half (3,081 patients); NIH failure-detection results refer to the evaluation half.

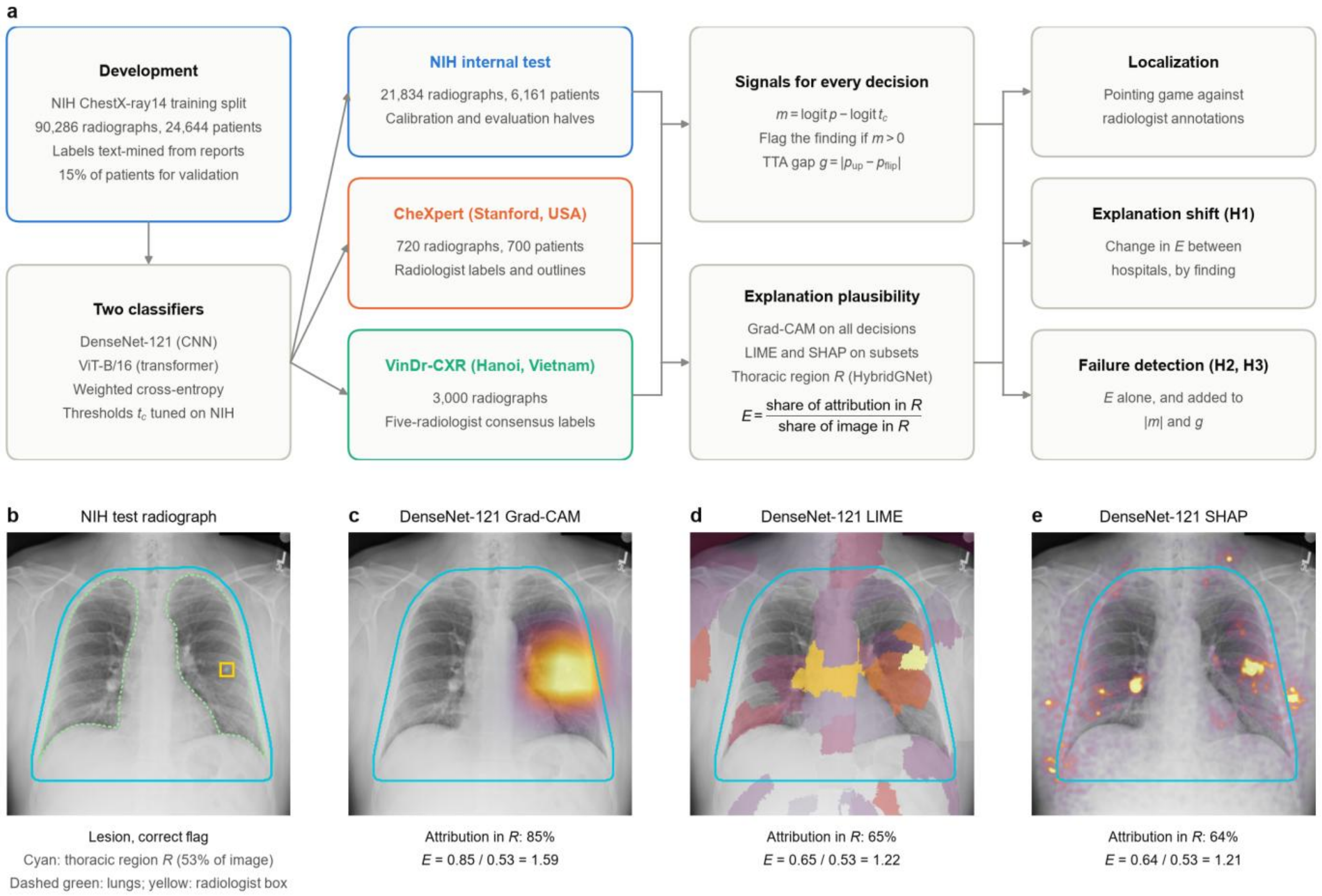

**Fig. 1 | Study design and anatomical enrichment.** **a**, DenseNet-121 and ViT-B/16 were trained on NIH ChestX-ray14 and evaluated at the training hospital (NIH internal test set, split by patient into a calibration half and an evaluation half) and at two external sites (CheXpert radiographs from Stanford Hospital, USA; VinDr-CXR radiographs from two hospitals in Hanoi, Vietnam). For every decision we recorded the margin $m$ to the NIH-tuned threshold $t_c$, the test-time augmentation (TTA) gap $g$ and, from its explanation, the anatomical enrichment $E$. **b**–**e**, A correct DenseNet-121 lesion flag on an NIH test radiograph, chosen among correct flags with a radiologist box as the one whose Grad-CAM enrichment is closest to the median of all correct NIH flags. **b**, Thoracic region $R$ (cyan; convex hull of the HybridGNet lung and heart masks, extended into the costophrenic recesses and dilated), lungs (dashed green) and radiologist box (yellow). **c**–**e**, Positive Grad-CAM (**c**), LIME (**d**) and SHAP (**e**) attribution, scaled between the 1st and 99.5th percentiles. $E$ divides the share of attribution inside $R$ by the share of the image that $R$ covers, so uniform attribution gives $E = 1$. Only NIH radiographs are shown in this paper.

## Ranking transferred to new hospitals but operating points did not

Discrimination held up away from NIH (Fig. 2a). On the eight findings shared with CheXpert, macro AUROC fell from 0.857 at NIH to 0.824 (0.794–0.853) for DenseNet-121 and from 0.844 to 0.803 (0.766–0.838) for ViT-B/16. On the ten findings evaluable at VinDr-CXR it rose, from 0.829 to 0.893 (0.881–0.905) and from 0.817 to 0.886 (0.874–0.895), consistent with cleaner consensus labels than the report-mined NIH labels[31]. Thresholds tuned on NIH validation data transferred far worse than rankings (Fig. 2b). At CheXpert, the sensitivity of DenseNet-121 fell from 0.424 to 0.008 for edema and from 0.409 to 0.136 for cardiomegaly, although both findings were still ranked with AUROCs above 0.82, and ViT-B/16 detected no edema at all. Sensitivity also rose for some findings, for example from 0.495 to 0.700 for ViT-B/16 atelectasis at CheXpert and from 0.202 to 0.447 for ViT-B/16 fibrosis at VinDr-CXR. Each model therefore made a different mix of errors at each hospital, which is when a per-case warning would be most useful.

## Saliency maps rarely pointed where radiologists did

On the 520 CheXpert findings with a radiologist benchmark, the Grad-CAM peak fell inside the reference segmentation for 0.575 (0.529–0.619) of DenseNet-121 findings and 0.381 (0.336–0.425) of ViT-B/16 findings, against 0.888 (0.862–0.915) for radiologists marking the most representative point (Fig. 3a). Across all annotated findings, DenseNet-121 hit rates were 0.487 at NIH ($n$ = 191), 0.541 at CheXpert ($n$ = 827) and 0.327 at VinDr-CXR ($n$ = 1,243); ViT-B/16 reached 0.188, 0.368 and 0.224 (Fig. 3b). Localization depended strongly on the finding: DenseNet-121 found cardiomegaly reliably (0.964 at NIH) but pleural thickening at VinDr-CXR almost never (0.053), and ViT-B/16 never placed its peak on an NIH infiltration or pneumothorax (Extended Data Fig. 1). For DenseNet-121, flagged findings were localized better than annotated findings in general (CheXpert 0.687 versus 0.541), as expected if correct flags rest on lesion evidence; for ViT-B/16 they were localized worse (0.151 versus 0.368). Grad-CAM was a weak explainer for the transformer. Its enrichment correlated poorly with that of LIME and SHAP (Spearman ρ 0.079–0.222), whereas LIME and SHAP agreed with each other (ρ 0.508–0.615; Extended Data Fig. 2a), so we treat LIME and SHAP as the reference explainers for ViT-B/16.

**Table 1 | Test cohorts, labels and harmonized findings.**

| | NIH ChestX-ray14 (internal) | CheXpert (external) | VinDr-CXR (external) |
|---|---|---|---|
| Hospital | NIH Clinical Center, USA | Stanford Hospital, USA | Hospital 108 and Hanoi Medical University Hospital, Vietnam |
| Radiographs (patients) | 21,834 (6,161) | 720 (700) | 3,000 (3,000)[a] |
| Labels | Mined from radiology reports | Consensus of 3 (validation) or 5 (test) radiologists | Consensus of 5 radiologists |
| View | PA 61.6%, AP 38.4% | AP 83.7%, PA 16.3%[b] | PA only[c] |
| Age, median (IQR), years | 48 (34–58) | 62 (48–75)[b] | 52 (35–63)[d] |
| Male | 57.1% | 53.5%[b] | 56.0%[d] |
| Reference localizations | 191 boxes on 171 radiographs | 827 segmentations on 426 radiographs; 520 benchmarked by radiologists | 1,660 boxes on 774 radiographs (1,243 radiograph–finding pairs) |
| Decisions per model | 262,008 | 5,760 | 36,000 |
| ***Positive radiographs by harmonized finding*** | | | |
| Atelectasis | 2,251 | 233 | 86 |
| Cardiomegaly | 540 | 220 | 309 |
| Consolidation | 847 | 61 | 96 |
| Edema | 484 | 121 | 0[e] |
| Effusion | 2,618 | 169 | 111 |
| Emphysema | 502 | – | 3[e] |
| Fibrosis | 332 | – | 217 |
| Infiltration | 3,893 | – | 58 |
| Lesion[f] | 2,197 | 10 | 176 |
| Pleural thickening | 692 | – | 169 |
| Pneumonia | 276 | 19 | 246 |
| Pneumothorax | 940 | 16 | 18 |
| ***Flags (findings called positive), DenseNet-121 / ViT-B/16*** | | | |
| Flags | 19,862 / 20,214 | 480 / 679 | 1,160 / 1,502 |
| Flag precision | 0.358 / 0.343 | 0.529 / 0.514 | 0.434 / 0.370 |

[a]No patient identifiers; each radiograph was treated as one patient. [b]Recorded for the 202 validation radiographs only. [c]Adult postero-anterior radiographs only, as stated by the data provider. [d]From DICOM headers where recorded (age for 468 and sex for 1,254 of 3,000 radiographs). [e]Excluded from per-finding transfer metrics (fewer than ten positives). [f]NIH mass or nodule; CheXpert lung lesion; VinDr-CXR nodule/mass. A dash means that the finding is not labeled at that site. AP, antero-posterior; PA, postero-anterior; IQR, interquartile range.

## Explanations shifted between hospitals, but not in a consistent direction (H1)

DenseNet-121 attribution for flagged findings was concentrated on the chest at every site, with median enrichment of 1.55–1.67 for correct and wrong flags and fewer than 3% of flags below the uniform level of 1 (Fig. 4a). ViT-B/16 Grad-CAM was close to uniform (medians 0.98–1.21; 21–51% of flags below 1). Pooled over findings, DenseNet-121 enrichment did not shift (CheXpert −0.010, 95% CI −0.042 to 0.028, $P = 0.58$; VinDr-CXR −0.011, −0.049 to 0.032, $P = 0.31$). Within findings it did (Fig. 4b and Extended Data Table 2). After Holm correction, five of six findings were less concentrated at CheXpert, including cardiomegaly (−0.230) and pneumothorax (−0.193), and four of ten at VinDr-CXR, while none became more concentrated; the pooled null reflects different case mixes, not stable explanations. H1 was therefore supported within findings for DenseNet-121. For ViT-B/16 the shifts went both ways. Pooled enrichment rose at CheXpert (+0.203, 0.172 to 0.234; $P < 10^{-16}$), and at VinDr-CXR four findings became less concentrated and two more concentrated (Holm-adjusted $P \leq 0.012$), so H1 as stated was not supported. Explanation statistics thus respond to a change of hospital, which population-level monitoring could detect[26], but the direction depends on the finding and the architecture, so a shift alone does not say whether decisions have become less reliable.

## Explanation plausibility did not identify wrong flags (H2)

If plausible explanations accompanied correct decisions, enrichment would separate true from false positives. We measured this with the AUROC for failure detection, AUROC$_f$, which equals 0.5 for an uninformative score and is higher when correct decisions score above wrong ones (Methods). At NIH, enrichment separated them weakly at best (Fig. 5a): AUROC$_f$ was 0.484 (0.468–0.502) for DenseNet-121 and 0.524 (0.509–0.539) for ViT-B/16, against 0.646 (0.632–0.662) and 0.660 (0.647–0.673) for the absolute decision margin. At CheXpert it stayed at chance for both models (0.487 and 0.536, with intervals including 0.5). At VinDr-CXR, DenseNet-121 enrichment became informative (0.618, 0.586–0.649), although it remained below the margin (0.685), whereas ViT-B/16 enrichment did not (0.522). The margin was the best single score in five of the six model–site combinations (AUROC$_f$ 0.646–0.685 for flags), the exception being CheXpert, where test-time augmentation agreement performed similarly for DenseNet-121 (0.616 versus 0.607).

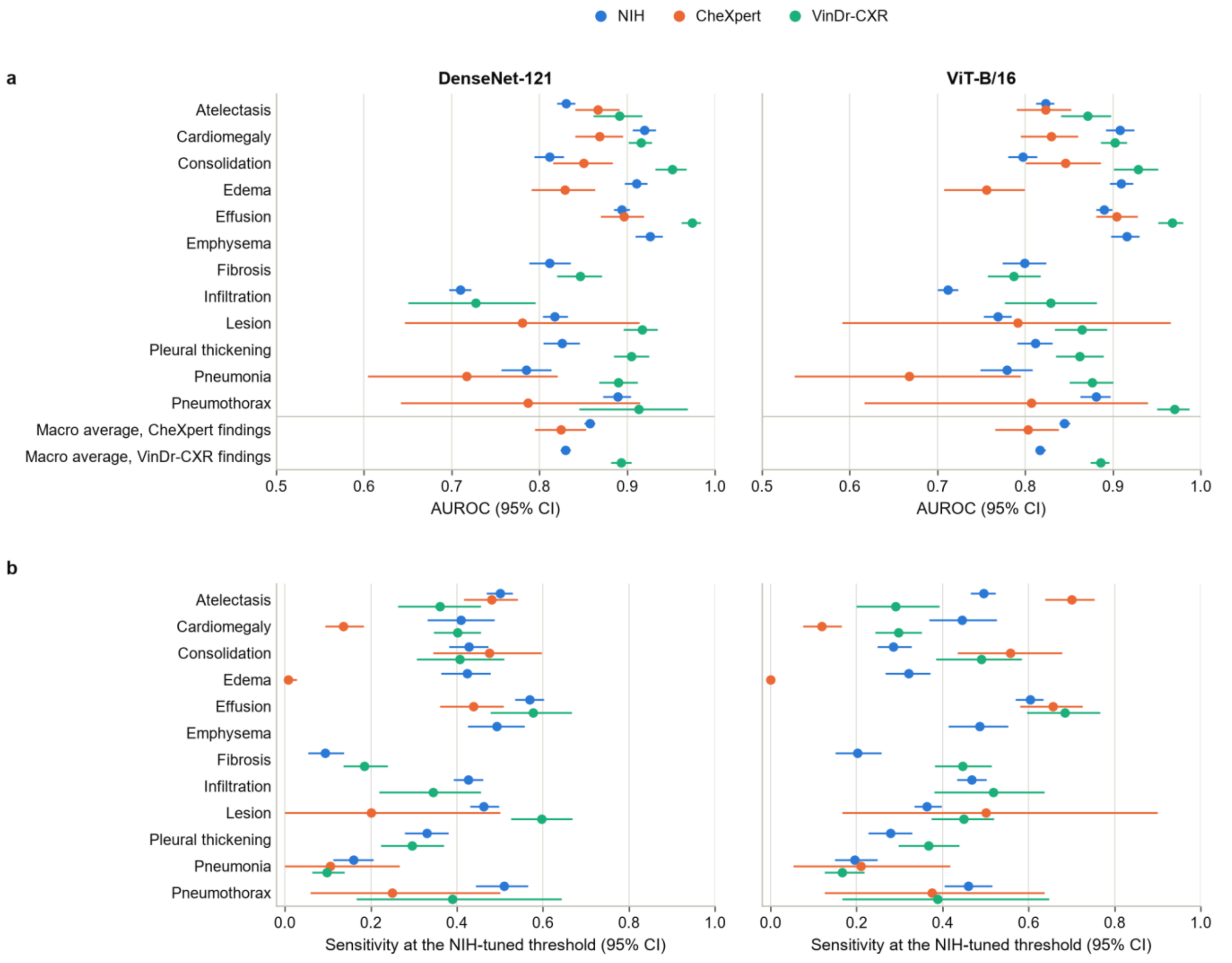

**Fig. 2 | Discrimination transferred to new hospitals but operating points did not. a**, Area under the receiver operating characteristic curve (AUROC) by finding at NIH (blue), CheXpert (orange) and VinDr-CXR (green) for DenseNet-121 (left) and ViT-B/16 (right). Below the line are macro averages over the findings shared with each external site, with the NIH value computed over the same findings. **b**, Sensitivity at the thresholds tuned on NIH validation data. Points show estimates and bars show 95% confidence intervals from 1,000 bootstrap resamples of patients (VinDr-CXR: radiographs). Findings with fewer than ten positives at a site are not shown (VinDr-CXR edema and emphysema), and CheXpert labels cover eight findings. Positives per finding and site are given in Table 1.

The VinDr-CXR result is not an artifact of Grad-CAM. On subsets of up to 500 radiographs per site, LIME and SHAP enrichment for DenseNet-121 moved the same way, from at or below chance at NIH (0.496 and 0.454) to above chance at VinDr-CXR (0.600 and 0.556), and LIME was below chance at CheXpert (0.411, 0.359–0.460; Fig. 5c). For ViT-B/16, LIME and SHAP enrichment stayed between 0.470 and 0.544 at every site.

Within findings the signal was weaker still. In finding–site cells with at least 20 correct and 20 wrong flags, enrichment exceeded chance in 3 of 20 cells for DenseNet-121 and in none of 23 for ViT-B/16, for which it fell below chance in 6, whereas the margin exceeded chance in 14 and 18 cells (Extended Data Fig. 3a). The median cell AUROC$_f$ of DenseNet-121 enrichment was 0.51 at NIH and 0.53 at VinDr-CXR, against pooled values of 0.484 and 0.618, so much of the pooled difference between these hospitals reflects which findings were flagged and how often each was wrong rather than a change within findings. H2, which was specified for pooled flags, was supported only for DenseNet-121 at VinDr-CXR and, weakly, for ViT-B/16 at NIH, where enrichment exceeded chance for no single finding.

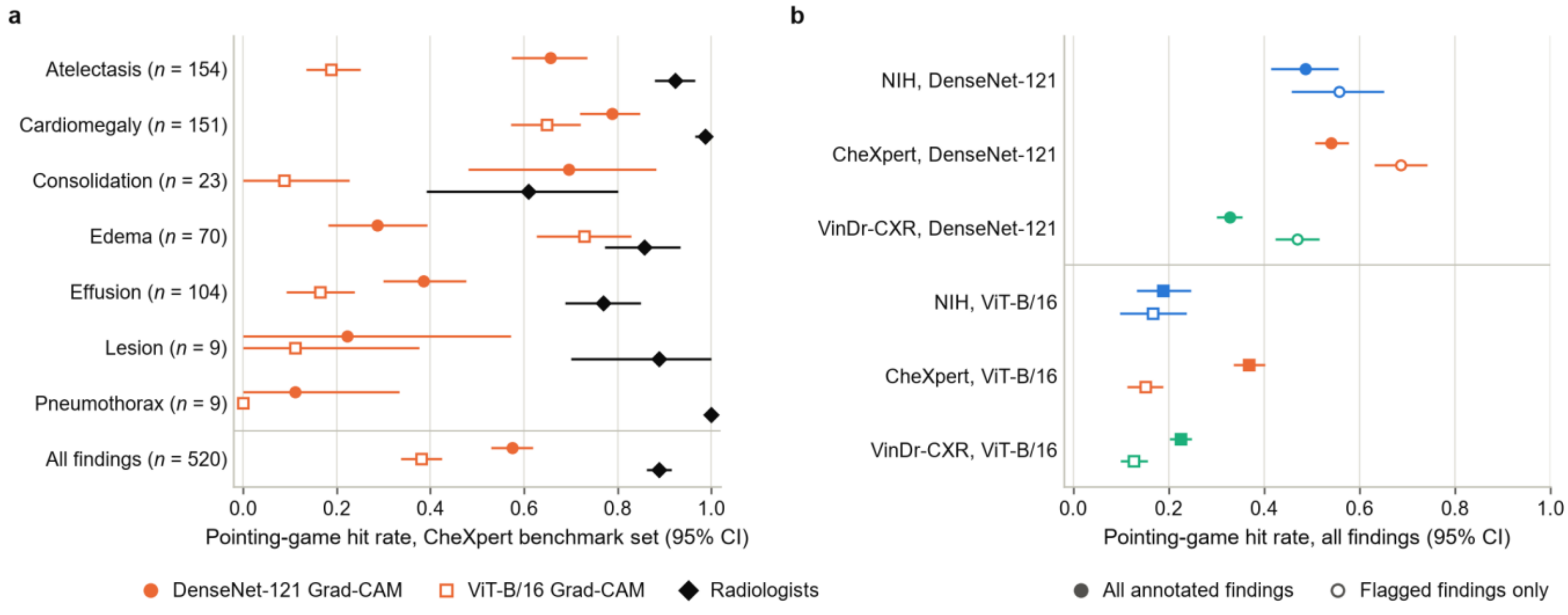

**Fig. 3 | Saliency maps rarely pointed where radiologists did. a**, Pointing-game hit rate on the CheXpert benchmark set, on which radiologists also localized each finding: DenseNet-121 Grad-CAM (filled circles), ViT-B/16 Grad-CAM (open squares) and the radiologists' most representative points (diamonds); *n* is the number of benchmarked findings. **b**, Hit rate over all annotated findings (filled) and over the findings that each model flagged (open), by site. A hit is scored when the peak of the map falls inside the reference segmentation (CheXpert) or box (NIH and VinDr-CXR). Bars show 95% confidence intervals (patient bootstrap; VinDr-CXR: radiographs). Annotated findings: NIH 191, CheXpert 827 and VinDr-CXR 1,243.

## Adding explanations to confidence did not help and could hurt (H3)

The pre-specified test combined the margin and test-time augmentation agreement in a logistic model fitted on the NIH calibration half, with and without enrichment, and compared the two on held-out decisions (Fig. 5b and Extended Data Table 3). Adding enrichment never improved failure detection meaningfully. For flagged findings, the change in $AUROC_f$ ($\Delta AUROC_f$) was +0.0022 (−0.0011 to 0.0052) at NIH, −0.0087 (−0.0205 to 0.0044) at CheXpert and −0.0233 (−0.0315 to −0.0155) at VinDr-CXR for DenseNet-121, and +0.0002 (−0.0023 to 0.0024), +0.0048 (−0.0004 to 0.0104) and +0.0017 (−0.0037 to 0.0068) for ViT-B/16. The largest upper confidence bound was 0.0104, so a gain above 0.011 is excluded for both models at every site, and the same holds for unflagged findings. The small reductions in the area under the risk–coverage curve (AURC) for ViT-B/16 that reached significance were at most 0.004, on AURCs of 0.35–0.52. H3 was not supported.

At VinDr-CXR, enrichment made DenseNet-121 worse (AURC 0.427 with enrichment versus 0.415 without; Fig. 5e and Extended Data Fig. 4), and the reason matters beyond this study. At NIH, conditional on the margin and test-time augmentation agreement, more concentrated maps were slightly more often wrong, so the combination fitted there gave enrichment a negative standardized weight (−0.071; Fig. 5d). At VinDr-CXR the relation ran the other way, with enrichment alone informative in the positive direction, so the transferred weight pointed the wrong way. When we refitted the combination within each site by grouped five-fold cross-validation, an exploratory analysis, enrichment received a positive weight at VinDr-CXR (+0.348) but still did not improve detection ($\Delta AUROC_f$ +0.0091, −0.0080 to 0.0240), and no refitted combination improved the detection of wrong flags at any site for either model (Fig. 5b,d). The relation between explanation plausibility and correctness therefore depends on the deployment site, and even where it was estimated with local labels it added nothing for flags. In practical terms, keeping the half of DenseNet-121 flags with the largest margin lowered the false-flag rate at VinDr-CXR from 0.566 to 0.421, whereas keeping the half with the highest enrichment gave 0.503, and at NIH selecting by enrichment was worse than keeping every flag (0.648 versus 0.636; Extended Data Fig. 3b).

## Missed findings, robustness and sanity checks

Among unflagged findings, where a failure is a missed finding, DenseNet-121 enrichment was informative in reverse: missed findings had more concentrated class evidence than correct negatives, giving $AUROC_f$ of 0.351 at NIH, 0.300 at CheXpert and 0.221 at VinDr-CXR (Extended Data Fig. 5a). This is the signature of a near miss, strong but subthreshold evidence, which the margin already captures ($AUROC_f$ 0.697–0.842). Accordingly, for DenseNet-121 the enrichment of unflagged findings was correlated with the absolute margin (Spearman ρ −0.36 to −0.50), whereas for flagged findings the two were nearly independent for both models ($|\rho| \leq 0.15$; Extended Data Fig. 5c), so the null result for flags does not arise from redundancy. With the combination fitted at NIH, adding enrichment changed $AUROC_f$ for unflagged findings by at most +0.0018 for DenseNet-121 and +0.0076 (0.0057 to 0.0094) for ViT-B/16. Refitting within each site recovered gains for missed findings at CheXpert (DenseNet-121 +0.030, 0.020 to 0.041; ViT-B/16 +0.052, 0.028 to 0.078) and for DenseNet-121 at VinDr-CXR (+0.024, 0.018 to 0.029), each time with a strongly negative weight on enrichment that was absent at NIH (DenseNet-121: −0.50 at CheXpert and −0.71 at VinDr-CXR versus −0.01; Extended Data Fig. 5b), so this information, too, was specific to the site.

Both architectures passed the model-randomization test[13]. Re-initializing the classification head decorrelated Grad-CAM maps from the trained maps (mean Spearman correlation 0.030 for DenseNet-121 and −0.125 for ViT-B/16), and re-initializing the last block as well kept them decorrelated (−0.006 and 0.154; Extended Data Fig. 2b), so the maps depend on what the models learned. The explainers nonetheless disagreed with one another. For DenseNet-121, Grad-CAM and SHAP enrichment correlated well (ρ 0.657–0.792) but LIME correlated weakly with both (ρ 0.063–0.454; Extended Data Fig. 2a). Typical correct and wrong flags at NIH show the same pattern (Extended Data Fig. 6).

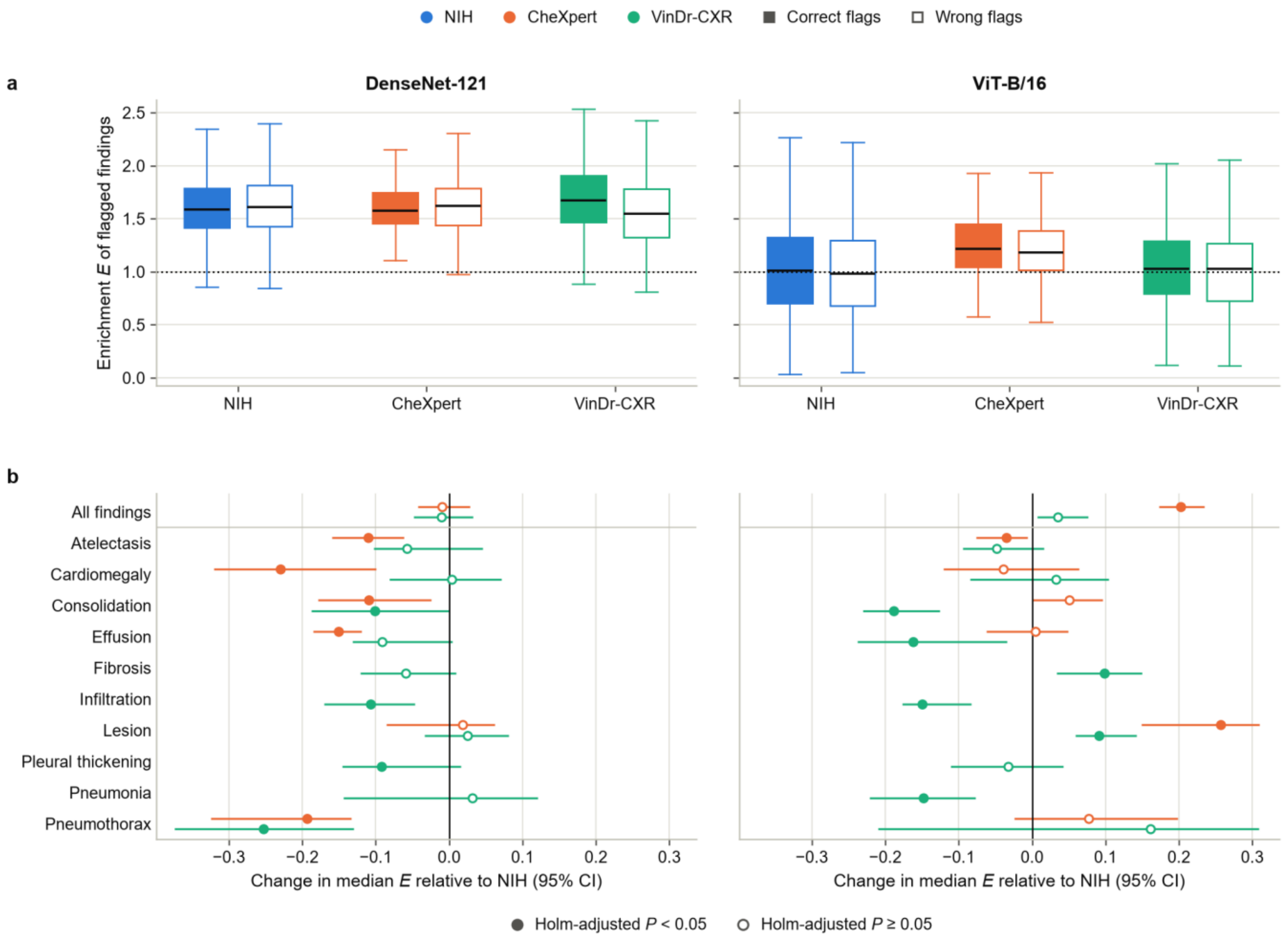

**Fig. 4 | Explanations shifted between hospitals, but not in a consistent direction.** **a**, Grad-CAM enrichment $E$ of correct (filled) and wrong (open) flags at each site. Boxes show medians and interquartile ranges, whiskers extend to 1.5 times the interquartile range, outliers are not drawn and the dotted line marks $E = 1$ (uniform attribution). Flags with usable masks: DenseNet-121 19,537, 480 and 1,160; ViT-B/16 19,860, 679 and 1,502 (NIH, CheXpert and VinDr-CXR). **b**, Change in the median $E$ of flagged findings at CheXpert (orange) and VinDr-CXR (green) relative to NIH, pooled and for each finding with at least 20 flags at both sites, with 95% bootstrap confidence intervals (patients resampled within each site). Filled markers indicate a Holm-adjusted two-sided Mann–Whitney $P < 0.05$ within each model and site (pooled row: unadjusted $P$).

## Discussion

Across two architectures, three sites in two countries and three explanation methods, the anatomical plausibility of an explanation did not tell us whether the decision it accompanied was correct, and combined with the model's own confidence before deployment it never added more than a negligible amount. The one setting in which plausibility carried clear information about flags, DenseNet-121 at VinDr-CXR, is the most instructive. There, plausibility predicted correctness in the opposite direction from the training hospital, so a check calibrated before deployment made failure detection worse after deployment, and recalibrating it with local labels did not make it useful. LIME and SHAP reproduced the change of direction, so it is a property of the model and the hospital rather than of Grad-CAM.

For networks whose head is global average pooling followed by a linear layer, the link between explanation and decision is exact. Grad-CAM reduces to the positive part of the class activation map, and the full map, summed over the image, is proportional to the class score minus its bias (equation (2) in Methods). The map is therefore not an independent measurement of the decision but a spatial rearrangement of the evidence that the score already summarizes. Enrichment depends on where that evidence lies rather than on how much there is, so it is not redundant by construction, and for flags it was almost uncorrelated with the margin. Yet where the evidence lay carried little information about correctness: a confident false positive is produced by strong evidence for the finding, and that evidence sits in the chest just as it does for a true positive. The vision transformer has no pooling head and its Grad-CAM maps were far less spatially specific, yet it gave the same answer with all three explainers, so the conclusion does not rest on the decomposition.

Why the relation reversed is less certain. Within findings, enrichment was close to chance at both NIH and VinDr-CXR, so much of the change in the pooled relation probably reflects case mix: which findings were flagged at each hospital, how concentrated their evidence typically is and how often each was wrong. Label quality differs as well. NIH labels were mined from reports and are noisy[31], so some NIH false positives are probably true findings that the labels missed, and such flags will carry lesion-like evidence; the VinDr-CXR labels are five-radiologist consensus. Acquisition differs too: VinDr-CXR contains adult postero-anterior radiographs only, whereas almost two in five NIH test radiographs and most CheXpert radiographs with recorded views are antero-posterior. Whatever the cause, the practical consequence is the same. A monitor applies one rule to every finding, so its behavior depends on the mix of findings and errors at the deployment site, which is not known in advance.

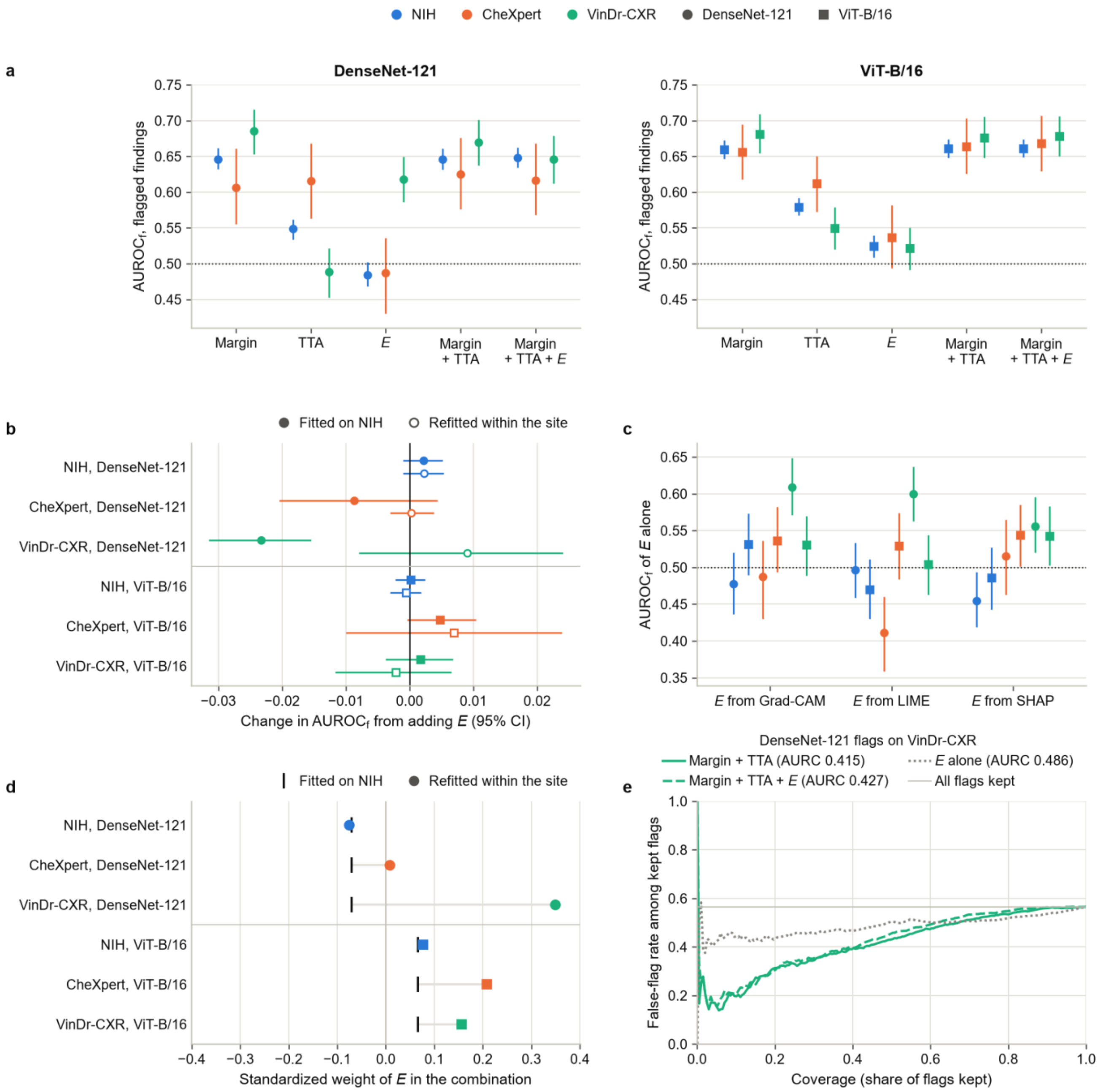

**Fig. 5 | Failure detection with and without explanation plausibility.** **a**, AUROCf for flagged findings (0.5, uninformative; higher values mean that correct flags score above wrong ones) of the absolute margin, TTA agreement, enrichment *E* and logistic combinations of the margin and TTA agreement without and with *E*, at NIH (evaluation half), CheXpert and VinDr-CXR. **b**, Change in AUROCf from adding *E* to the margin–TTA combination, fitted on the NIH calibration half (filled; pre-specified) or refitted within each site by grouped five-fold cross-validation (open; exploratory). **c**, AUROCf of *E* alone computed from Grad-CAM, LIME and SHAP on the cross-method subsets (up to 500 radiographs with flags per site). **d**, Standardized weight of *E* in the combination fitted on NIH (vertical tick) and refitted within each site (marker). **e**, Risk–coverage curves for DenseNet-121 flags at VinDr-CXR, showing the false-flag rate among the flags kept when flags are retained in decreasing order of each score; the area under each curve (AURC) is given in the legend. Circles, DenseNet-121; squares, ViT-B/16. Bars show 95% confidence intervals from 1,000 bootstrap resamples of patients (VinDr-CXR: radiographs). Flags (wrong flags) for DenseNet-121 and ViT-B/16: NIH evaluation half 9,296 (5,908) and 9,603 (6,267); CheXpert 480 (226) and 679 (330); VinDr-CXR 1,160 (656) and 1,502 (946).

For deployment, our results separate safeguards that worked from one that did not. Operating points did not transfer: with NIH-tuned thresholds both models missed almost all edema at CheXpert, although they still ranked it reasonably well (AUROC 0.83 and 0.76), so thresholds need to be re-tuned on a local sample before use. Confidence was the most reliable error signal at every site, although modest ($AUROC_f$ 0.61–0.69 for flags): keeping the more confident half of flags removed 6–14 percentage points of false flags at every site, which makes it the natural basis for monitoring. A heatmap that looks right should not reassure anyone that a finding is correct: in our data, wrong flags were usually as anatomically plausible as correct ones. This agrees with the argument that post hoc explanations cannot provide patient-level assurance[16], with clinical evidence that explanations do not protect readers from a misleading model[15], and with the argument that current post hoc interpretability methods may not be the way to make AI trustworthy, because they capture correlations between inputs and outputs rather than how a model reaches its decisions[32].

Our findings extend evidence that saliency maps localize worse than radiologists[10] and fail trust criteria[11,12] from localization to error detection. Zain et al. concluded from VinDr-CXR alone that overlap between saliency and annotated boxes should not be taken as evidence of diagnostic correctness[24]; we observed the same at three sites, with two architectures, three explainers and confidence baselines, together with a site-dependent reversal that a single-site study cannot reveal. Within-dataset work that combined explanation quality with confidence reported marginal gains[25], and failure-detection benchmarks found that no confidence score consistently outperforms simple baselines[20,21]; our bounded null, which excludes gains above 0.011 $AUROC_f$ for a detector calibrated before deployment and above 0.024 for flags even after local recalibration, quantifies how little room explanations leave. Mamun et al. reported that attribution inside the lungs stayed largely stable under external transfer while calibration degraded[22]; we see the same dissociation between what explanations show and how reliable decisions are.

The study has limitations. NIH labels are report-mined and noisy[31], which affects internal failure counts, although both external sites used radiologist-consensus labels. Some external findings had few positives (CheXpert lesion 10, pneumothorax 16), so their per-finding estimates are imprecise. We studied one explanation summary; statistics other than spatial concentration, such as disagreement between explanation methods, which was large, might behave differently. For the transformer we did not evaluate transformer-specific attribution methods[33,34]. In lung CT classification, Grad-CAM was less faithful for vision transformers than for convolutional networks[35], and rankings of attribution methods established on convolutional networks do not reliably transfer to transformers[36], which is why we used LIME and SHAP as reference explainers for ViT-B/16. LIME and SHAP were computed on subsets for reasons of cost. The site-refitted, per-finding, correlation and selective-risk analyses were added after the pre-specified analysis and are exploratory, and most per-finding cells were small, with intervals about 0.2 wide. Small differences in floating-point arithmetic between the training and analysis environments changed 0.18% of ViT-B/16 decisions (Methods); all decisions, margins and maps come from the recomputed outputs and are therefore mutually consistent. Finally, we did not run a reader study, so we cannot say how radiologists would use heatmaps at a new hospital.

Two directions follow. If explanation-based checks are to be used at all, their relation to correctness must be estimated with labeled cases from each deployment site; in our data even this helped only with missed findings, and such cases are better spent re-tuning thresholds and confidence scores. Beyond heatmaps, natural-language explanations can state the reasons for a decision in a form that users can read and question[37], and report generators that use feedback from their own draft reports to refine their abnormality predictions[38] point to explanations that are checked rather than merely displayed. Whether such explanations can detect errors at a new hospital should be tested in the same way, against the model's confidence and across sites.

# Methods

### Data and ethics

We used four public resources under their providers' terms: NIH ChestX-ray14 (112,120 frontal radiographs from 30,805 patients at the NIH Clinical Center, each labeled for 14 findings mined from the radiology report)[1]; the CheXpert validation and test radiographs released with CheXlocalize (Stanford Hospital)[3,10]; the VinDr-CXR test set (Hospital 108 and Hanoi Medical University Hospital, Vietnam), version 1.0.0 on PhysioNet[30,39,40]; and the CheXmask database of anatomical masks, version 1.0.0 on PhysioNet[41–43]. All data were de-identified by their providers before release. The Stanford University Institutional Review Board reviewed the collection of CheXpert and waived informed consent because the data were retrospective and fully anonymized[10]. The institutional review boards of Hanoi Medical University Hospital and Hospital 108 approved the collection of VinDr-CXR and waived informed consent because the retrospective study did not affect patient care[30]. VinDr-CXR was accessed under the PhysioNet Credentialed Health Data Use Agreement and processed only on the investigators' own computer; no radiograph from VinDr-CXR or CheXpert is reproduced in this paper. Our analysis used only these de-identified data, with no access to identifying information and no contact with patients. Under the guidelines of the Nanyang Technological University Institutional Review Board, secondary research on such anonymized data is not considered human subjects research, so no further approval was sought. Reporting follows the Checklist for Artificial Intelligence in Medical Imaging[44].

### Cohorts and label harmonization

NIH ChestX-ray14 radiographs were assigned to training and test sets by patient, with 20% of patients held out. Because rare findings make random splits uneven, we generated 50 patient-grouped splits (seeds 0–49) and kept the one whose training and test sets differed least in the prevalence of the 14 NIH findings, measured as half the sum of the absolute differences in prevalence (0.0094, against 0.0296 for the most uneven candidate). The selected split contains 90,286 training radiographs from 24,644 patients and 21,834 test radiographs from 6,161 patients, with no patient in both. A further 15% of training patients formed a validation set (76,491 training and 13,795 validation radiographs).

The CheXpert validation set (234 radiographs from 200 patients, labeled by the consensus of three radiologists) and test set (668 radiographs from 500 patients, labeled by the consensus of five radiologists) were restricted to frontal radiographs, giving 720 radiographs from 700 patients[3,10]. The VinDr-CXR test set contains 3,000 postero-anterior radiographs of adults, each labeled by the consensus of five radiologists[30]; it carries no patient identifiers, so each radiograph was treated as one patient.

Twelve findings were harmonized across the three label sets (Table 1): atelectasis, cardiomegaly, consolidation, edema, effusion (CheXpert and VinDr-CXR pleural effusion), emphysema, fibrosis (VinDr-CXR pulmonary fibrosis), infiltration, lesion (NIH mass or nodule; CheXpert lung lesion; VinDr-CXR nodule/mass), pleural thickening, pneumonia and pneumothorax. Hernia has no external counterpart and was excluded. CheXpert labels cover eight of the twelve findings. At VinDr-CXR, edema (no positives) and emphysema (three positives) were excluded from per-finding transfer metrics, which require at least ten positives.

Reference localizations were 191 radiologist bounding boxes on 171 NIH test radiographs[1]; 827 radiologist segmentations on 426 CheXpert radiographs for seven findings (pneumonia is not segmented), with a human benchmark of segmentations and most-representative points drawn by separate radiologists on the test set, 520 of which match a reference segmentation[10]; and 1,660 consensus boxes on 774 VinDr-CXR radiographs, forming 1,243 radiograph–finding pairs once boxes for the same finding were merged.

### Image preprocessing

All radiographs were converted to 8-bit grayscale and resized to 320 × 320 pixels with bicubic interpolation. VinDr-CXR DICOM files (1,033 compressed with lossless JPEG 2000 and 1,967 uncompressed; bit depth 10–16) were decoded by applying the modality lookup table and then the value-of-interest lookup table, inverting MONOCHROME1 images, and min–max scaling to 8 bits before resizing.

### Classifiers and training

DenseNet-121[28] was initialized with ImageNet weights; the grayscale input was replicated to three channels and normalized with ImageNet statistics. For a 320-pixel image the final dense block produces a 10 × 10 × 1,024 feature map, which is average-pooled, passed through dropout (rate 0.5) and mapped by a linear layer to 14 logits $z_c$ with $p_c = \sigma(z_c)$. ViT-B/16[29] was initialized from ImageNet-pretrained weights for 384-pixel inputs and run at 320 pixels, giving 20 × 20 patches and 401 tokens; position embeddings were resized from 24 × 24 to 20 × 20 by bicubic interpolation, inputs were scaled to $[-1, 1]$, and the class token was passed through dropout (rate 0.5) and a linear layer to 14 logits. Class imbalance was handled with a binary cross-entropy averaged over the 14 classes, in which the positive examples of each class were weighted by the ratio of negative to positive training radiographs for that class, clipped to the range 1–8; the cap of 8 applies to 12 of the 14 classes, and effusion and infiltration receive weights of 7.4 and 4.7.

Training had two stages. The classification head was first trained for two epochs with the backbone frozen (Adam, learning rate $10^{-3}$). The whole network was then fine-tuned for 12 epochs with AdamW[45] (weight decay 0.05, not applied to biases, normalization parameters or, for ViT-B/16, the class token and embeddings), with an exponential moving average of the weights (decay 0.999) and a learning rate that rose linearly over the first epoch from 1% of its peak and then followed a cosine decay to 2% of its peak. Peak learning rates and batch sizes were $10^{-4}$ and 32 for DenseNet-121 and $2.5 \times 10^{-5}$ and 8 for ViT-B/16. Training images were augmented with horizontal flips, brightness and contrast changes of up to 15%, rotations of up to 14.4°, zoom of up to 10% and translations of up to 7%. We kept the checkpoint with the highest validation macro AUROC (DenseNet-121, epoch 7, 0.846; ViT-B/16, epoch 6, 0.831), which in both cases outperformed the averaged weights. At inference, the probability of class $c$, $\bar{p}_c(x)$, was the average of $\sigma(z_c)$ over the radiograph $x$ and its horizontal mirror image $\tilde{x}$. Each threshold $t_c$ maximizes F1 on the validation set; to reduce its variance we took the median of the F1-optimal cut points over 25 bootstrap resamples of the validation data. As reference models, a random forest (300 trees, maximum depth 20) and a one-versus-rest linear support vector machine ($C = 1$, balanced class weights) were trained on the standardized 1,024-dimensional pooled DenseNet-121 features, with thresholds tuned in the same way (Extended Data Table 1).

### Decisions, margins and failures

A harmonized finding $f$ may correspond to several model classes $S_f$; for instance, a lesion is either a mass or a nodule. We define the decision margin

$$m_f(x) = \max_{c \in S_f} [\operatorname{logit} \bar{p}_c(x) - \operatorname{logit} t_c] \qquad (1)$$

and the decision $d_f(x) = \mathbb{1}[m_f(x) > 0]$. The class $c^*$ attaining the maximum in equation (1), the driver class, supplies the probability, the test-time augmentation disagreement $g_f(x) = |\sigma(z_{c^*}(x)) - \sigma(z_{c^*}(\tilde{x}))|$ and the explanation for the decision. A decision fails when $d_f(x) \neq y_f(x)$, where $y_f$ is the reference label. Flagged findings ($d_f = 1$) fail by being false positives and were the primary analysis; unflagged findings fail by being missed findings and were a secondary analysis.

### Explanations

For DenseNet-121, Grad-CAM[46] weights each channel $k$ of the final feature map $A \in \mathbb{R}^{H \times W \times K}$ by the spatial average of the gradient of the class score $z_c$ with respect to that channel, sums the weighted channels and keeps the positive part. The head is global average pooling followed by a linear layer with weights $w_{kc}$ and biases $b_c$ (dropout is inactive at inference), so this gradient equals $w_{kc}/(HW)$ at every location. Grad-CAM therefore reduces, as Selvaraju et al.[46] showed, to the class activation map of Zhou et al.[47], $M_c(u, v) = \sum_k w_{kc} A_k(u, v)$, with negative values set to zero and divided by $HW$. The factor $1/(HW)$ leaves every statistic in this paper unchanged, because the location of the peak, the Otsu threshold of the max-normalized map and the ratio in equation (4) are invariant to positive rescaling. All 14 maps therefore follow from one forward pass. On eight NIH radiographs, the class activation map matched gradient-based Grad-CAM multiplied by $HW$ to within $1.53 \times 10^{-5}$, against a peak map value of 31.25.

As Zhou et al.[47] noted, the class activation map, summed over all locations, is proportional to the class score minus its bias,

$$\sum_{u,v} M_c(u,v) = HW(z_c - b_c) \quad (2)$$

so the map is a spatial rearrangement of the evidence that the class score already summarizes, drawn from the same feature map and the same weights, rather than an independent measurement of the decision. The identity does not make enrichment (equation (4)) redundant, because enrichment depends on where attribution lies rather than on how much there is.

ViT-B/16 has no pooling head, so we applied Grad-CAM to the patch tokens entering the attention layer of the last transformer block, that is, the output $A \in \mathbb{R}^{400\times768}$ of its first layer normalization, with weights summed over the 400 patch tokens $u$:

$$L_c(u) = \mathrm{ReLU}\left(\sum_k \alpha_k^c A_{u,k}\right), \qquad \alpha_k^c = \sum_{u'} \frac{\partial z_c}{\partial A_{u',k}} \quad (3)$$

This forward pass reproduced the model's own predictions exactly (largest difference 0). Maps from the upright and mirrored radiograph were averaged after flipping back and bilinearly upsampled to 320 × 320 pixels; we write $\tilde{L}$ for the upsampled map.

LIME[48] fits a local linear surrogate over quickshift superpixels; we used 1,000 perturbations, replaced masked superpixels with zero and read each superpixel's attribution from the surrogate weight. SHAP values[49] were estimated with the expected-gradients estimator[50], using 50 random NIH test radiographs as background and 100 samples per explanation. Only positive attributions were used. Because LIME and SHAP are expensive, they were computed for the flagged findings on up to 500 randomly chosen radiographs with at least one flag per site, drawing NIH radiographs from the evaluation half only (DenseNet-121: 500 of 4,779 NIH, 311 of 311 CheXpert and 500 of 716 VinDr-CXR radiographs; ViT-B/16: 500 of 4,938, 405 of 405 and 500 of 857).

## Anatomical reference and thoracic region

Lung and heart masks come from HybridGNet[51], the landmark-based model used to build CheXmask[41]. For NIH radiographs we used the released CheXmask masks and discarded those whose estimated reverse-classification-accuracy Dice fell below 0.7, as the CheXmask authors recommend[42]; 21,564 of 21,834 (98.8%) passed. For consistency with our own image decoding, we segmented all CheXpert and VinDr-CXR radiographs with the released HybridGNet weights, following the authors' preprocessing (centered zero-padding to a square and resizing to 1,024 × 1,024 pixels), and kept masks that passed a plausibility check: both lungs and the heart present, the lungs side by side and lung area between 10% and 70% of the image. All 720 CheXpert and 2,999 of 3,000 VinDr-CXR masks passed. Where CheXmask also provides masks, agreement was near perfect, with lung Dice of 0.995 (5th percentile 0.988) on 200 CheXpert radiographs, 0.997 (0.996) on all 3,000 VinDr-CXR radiographs and 0.997 (0.996) on 300 random NIH radiographs, and heart Dice of 0.995, 0.997 and 0.998 (Extended Data Table 4).

The thoracic region $R$ is the convex hull of the lung and heart masks, extended by two points at the lateral limits of the lungs placed 4% of the image height below the lowest lung pixel, and dilated with an elliptical structuring element of radius 2.5% of the image size (8 pixels at 320; Fig. 1b). The extension closes the costophrenic recesses, where effusion and basal atelectasis lie and where a plain hull would cut across. Radiologist annotations lie almost entirely inside $R$: on average 99.0% of the pixels of the 827 CheXpert segmentations, 98.0% of the 1,660 VinDr-CXR boxes and 96.3% of the 188 NIH boxes on radiographs with usable masks (medians 100%); the lower NIH figure reflects boxes that extend over the chest wall.

## Anatomical enrichment

For an explanation map $\tilde{L} \geq 0$ on the image domain $\Omega$, anatomical enrichment is

$$E = \frac{\dfrac{\sum_{u\in R} \tilde{L}(u)}{\sum_{u\in \Omega} \tilde{L}(u)}}{\dfrac{|R|}{|\Omega|}} \quad (4)$$

The numerator is the share of attribution that falls in the thoracic region and the denominator is the share of the image that the region occupies. $E$ is invariant to rescaling of $\tilde{L}$, equals 1 when attribution is spread uniformly and lies between 0 and $|\Omega|/|R|$. The normalization matters across hospitals, which frame the chest differently and would otherwise produce different raw shares for the same behavior. Maps that are zero everywhere are assigned $E = 1$, and decisions on radiographs without a usable mask were excluded from analyses of $E$.

## Hypotheses and end points

Before analysis we specified three hypotheses, writing $\mathcal{D}_{\mathrm{int}}$ and $\mathcal{D}_{\mathrm{ext}}$ for the internal and external test distributions. H1: for flagged findings of the same type, $E$ is lower under $\mathcal{D}_{\mathrm{ext}}$ than under $\mathcal{D}_{\mathrm{int}}$. H2: among flagged findings, $\mathrm{AUROC_f}(E) > 0.5$, so that true positives have higher $E$ than false positives. H3: under $\mathcal{D}_{\mathrm{ext}}$, adding $E$ to a confidence score built from the decision margin and test-time augmentation increases $\mathrm{AUROC_f}$. The primary end point was the resulting change, $\Delta\mathrm{AUROC_f}$, for flagged findings at the external sites; $\Delta$AURC was secondary. The second external site (VinDr-CXR) and the second architecture (ViT-B/16) were analyzed with the same code and the same pre-specified hypotheses and end points. The analyses described under 'Supplementary analyses' were added afterwards and are exploratory.

### Failure-detection scores and metrics

A confidence score κ should rank correct decisions above failures. We measured how well it does so with $AUROC_f$, the area under the ROC curve for separating correct decisions from failures by κ (ties counted as one half), and with the area under the risk–coverage curve[17,19] (AURC), the failure rate among the $r$ highest-scoring decisions averaged over $r = 1, \ldots, n$, where $n$ is the number of decisions. Higher $AUROC_f$ and lower AURC are better, and $AUROC_f = 0.5$ means the score carries no information about correctness. The candidate scores were the absolute margin $|m|$, which plays the role of the softmax confidence baseline[18], the negative test-time augmentation disagreement $-g$[52], enrichment $E$ and two logistic combinations,

$$s_{\mathrm{MT}} = \sigma(\beta_0 + \beta_1 \hat{m} + \beta_2 \hat{g}),$$
$$s_{\mathrm{MTE}} = \sigma(\gamma_0 + \gamma_1 \hat{m} + \gamma_2 \hat{g} + \gamma_3 \hat{E}) \quad (5)$$

where σ is the logistic function and $\hat{m}$, $\hat{g}$ and $\hat{E}$ are $|m|$, $g$ and $E$ standardized to zero mean and unit variance. To keep the combinations honest, NIH test patients were split at random into halves (seed 42); the coefficients of $\ell_2$-regularized logistic regression (inverse regularization strength $C = 1$) and the standardization were fitted on the calibration half (3,080 patients), and every score was evaluated on the evaluation half (3,081 patients) and at the external sites. The effect of adding enrichment is $\Delta AUROC_f = AUROC_f(s_{\mathrm{MTE}}) - AUROC_f(s_{\mathrm{MT}})$, and ΔAURC is defined in the same way.

### Localization

For a finding with reference region $G$, a pointing-game hit[53] is scored when the peak of the map falls inside $G$, and the overlap is the intersection over union between $G$ and the region where the max-normalized map reaches its Otsu threshold[54]. Maps that are zero everywhere count as misses. For the radiologist benchmark, a hit is scored when the radiologist's most-representative point falls inside $G$, and the overlap compares the benchmark segmentation with the reference. We report all annotated findings and the subset that each model flagged; per-finding results require at least five annotations.

### Explanation shift

For H1 we compared the median enrichment of flagged findings at each external site with that at NIH, overall and for each finding with at least 20 flags at both sites, using the difference of medians with a percentile bootstrap that resamples patients independently within each site, and a two-sided Mann–Whitney $U$ test on the same values. Within each model and external site, $P$ values across the findings tested were adjusted with the Holm procedure[55]. Because the rank test compares whole distributions and treats flags as independent, whereas the bootstrap interval is for the difference of medians and resamples patients, the two can disagree for borderline differences, as for DenseNet-121 pleural thickening at VinDr-CXR (Fig. 4b and Extended Data Table 2).

### Sanity checks and agreement between explainers

Following Adebayo et al.[13], we re-initialized the classification head with Glorot-uniform weights and then, cumulatively, the last dense block (DenseNet-121) or the last transformer block (ViT-B/16), and computed the Spearman correlation between trained and randomized maps at each radiograph's top-scoring class for 200 random NIH test radiographs. Agreement between explainers was measured as the Spearman correlation of their enrichment values over the same flagged findings.

### Supplementary analyses

Four exploratory analyses used only the saved per-decision outputs. (1) Site-refitted combinations: within each evaluation partition (NIH evaluation half, CheXpert and VinDr-CXR) and scope, both combinations in equation (5) were refitted by five-fold cross-validation stratified by outcome and grouped by patient (StratifiedGroupKFold, seed 42), and $\Delta AUROC_f$ and ΔAURC were computed on the out-of-fold predictions; standardized coefficients were obtained from a fit on the whole partition. (2) Per-finding failure detection: $AUROC_f$ of $|m|$ and of $E$ for each finding and site with at least 20 correct and 20 wrong flags. (3) Redundancy: the Spearman correlation between $E$ and $|m|$ in each partition and scope. (4) Selective risk: the false-flag rate among the 50% of flags with the highest score, for each score. Intervals were obtained by the patient bootstrap described below.

### Statistics

Confidence intervals are 95% percentile intervals from 1,000 bootstrap resamples of patients (VinDr-CXR: radiographs; seed 42). Differences between scores were computed on the same resamples and are therefore paired. Per-finding AUROCs require at least ten positives, and failure-detection estimates at least 20 correct and 20 failed decisions. For the model-randomization test we report the mean correlation with the 2.5th and 97.5th percentiles across radiographs. $P$ values that evaluate to zero in double precision are reported as $P < 10^{-16}$.

### Software, hardware and numerical reproducibility

Experiments used Python 3.12 with TensorFlow 2.18, Keras 3.7 and KerasHub 0.21.1 for the classifiers and Grad-CAM, scikit-learn 1.9 for the reference models, logistic regression and metrics, SHAP 0.52, LIME 0.2.0.1, pydicom 3.0 with pylibjpeg for DICOM decoding, and PyTorch 2.14 with PyTorch Geometric 2.4.0 for HybridGNet, which ran on a central processing unit. DenseNet-121 was trained, and all analyses were run, on one NVIDIA GeForce RTX 4070 laptop GPU (8 GB) with mixed float16 precision. ViT-B/16 was trained on one NVIDIA A100 GPU on Google Colab with mixed bfloat16 precision and Keras 3.15.1, and analyzed locally with Keras 3.7. The two Keras versions compute bfloat16 attention logits differently (Keras 3.7 casts queries and keys to float32 before the product, whereas Keras 3.15.1 casts after it), so locally recomputed ViT-B/16 probabilities differed from the training-time outputs by a median of $8.5 \times 10^{-4}$ (maximum 0.190), changing 553 of 305,676 internal decisions (0.18%); DenseNet-121 outputs changed 7 of 305,676 decisions. NIH per-class AUROCs agreed to three decimals. Extended Data Table 1 reports the training-time outputs, and every other analysis used the locally recomputed outputs. All other random seeds were fixed at 42.

### Use of large language models

During the preparation of this manuscript, the authors used Claude (Opus 5.5, Anthropic) to edit and refine the manuscript text from the authors' drafts and to improve its grammar, sentence flow and clarity, as well as for coding assistance, debugging and refactoring on the figure-plotting and supplementary evaluation scripts. The research question, study design, core methodologies and experimental execution originated with the authors. After using this tool, the authors thoroughly reviewed, edited and verified all text, code and outputs to ensure technical accuracy, and take full responsibility for the integrity and content of this publication.

### Data availability

NIH ChestX-ray14 is available from the NIH Clinical Center at https://nihcc.app.box.com/v/ChestXray-NIHCC. The CheXpert validation and test radiographs with CheXlocalize annotations are available from the Stanford Center for Artificial Intelligence in Medicine and Imaging at https://stanfordaimi.azurewebsites.net/datasets/23c56a0d-15de-405b-87c8-99c30138950c under its research use agreement. VinDr-CXR is available on PhysioNet to credentialed users who sign its data use agreement (https://doi.org/10.13026/3akn-b287). The CheXmask database is openly available on PhysioNet (https://doi.org/10.13026/3705-zg36). No radiograph from CheXpert or VinDr-CXR is reproduced in this paper, and per-radiograph outputs derived from VinDr-CXR cannot be redistributed under its data use agreement. Aggregate results are provided with the code.

### Code availability

The code for data preparation, model training, anatomical segmentation, explanation analysis and figures is available at https://github.com/aashish-chapagai/CXR-Saliency-External-Validation. The version used in this paper is archived at Zenodo together with the trained DenseNet-121 and ViT-B/16 weights and their decision thresholds (https://doi.org/10.5281/zenodo.23618485)[56]. Code and models are released under the MIT license. HybridGNet code and weights are available from their authors at https://huggingface.co/spaces/ngaggion/Chest-x-ray-HybridGNet-Segmentation.

## References

1. Wang, X. *et al.* ChestX-Ray8: Hospital-Scale Chest X-Ray Database and Benchmarks on Weakly-Supervised Classification and Localization of Common Thorax Diseases. in *2017 IEEE Conference on Computer Vision and Pattern Recognition (CVPR)* 3462–3471 (IEEE, Honolulu, HI, 2017). doi:10.1109/CVPR.2017.369.
2. Rajpurkar, P. *et al.* Deep learning for chest radiograph diagnosis: A retrospective comparison of the CheXNeXt algorithm to practicing radiologists. *PLOS Med.* **15**, e1002686 (2018).
3. Irvin, J. *et al.* CheXpert: A Large Chest Radiograph Dataset with Uncertainty Labels and Expert Comparison. *Proc. AAAI Conf. Artif. Intell.* **33**, 590–597 (2019).
4. Zech, J. R. *et al.* Variable generalization performance of a deep learning model to detect pneumonia in chest radiographs: A cross-sectional study. *PLOS Med.* **15**, e1002683 (2018).
5. DeGrave, A. J., Janizek, J. D. & Lee, S.-I. AI for radiographic COVID-19 detection selects shortcuts over signal. *Nat. Mach. Intell.* **3**, 610–619 (2021).
6. Luo, L. *et al.* Rethinking Annotation Granularity for Overcoming Shortcuts in Deep Learning–based Radiograph Diagnosis: A Multicenter Study. *Radiol. Artif. Intell.* **4**, e210299 (2022).
7. Ong Ly, C. *et al.* Shortcut learning in medical AI hinders generalization: method for estimating AI model generalization without external data. *Npj Digit. Med.* **7**, 124 (2024).
8. Bungert, T. J., Kobelke, L. & Jäger, P. F. Understanding Silent Failures in Medical Image Classification. in *Medical Image Computing and Computer Assisted Intervention – MICCAI 2023* (eds Greenspan, H. et al.) vol. 14222 400–410 (Springer Nature Switzerland, Cham, 2023).
9. Lekadir, K. *et al.* FUTURE-AI: international consensus guideline for trustworthy and deployable artificial intelligence in healthcare. *BMJ* **388**, e081554 (2025).
10. Saporta, A. *et al.* Benchmarking saliency methods for chest X-ray interpretation. *Nat. Mach. Intell.* **4**, 867–878 (2022).
11. Arun, N. *et al.* Assessing the Trustworthiness of Saliency Maps for Localizing Abnormalities in Medical Imaging. *Radiol. Artif. Intell.* **3**, e200267 (2021).
12. Zhang, J. *et al.* Revisiting the Trustworthiness of Saliency Methods in Radiology AI. *Radiol. Artif. Intell.* **6**, e220221 (2024).
13. Adebayo, J. *et al.* Sanity Checks for Saliency Maps. in *Neural Information Processing Systems* 9505–9515 (2018).
14. Viviano, J. D., Simpson, B., Dutil, F., Bengio, Y. & Cohen, J. P. Saliency Is a Possible Red Herring When Diagnosing Poor Generalization. in *International Conference on Learning Representations* (2021).
15. Jabbour, S. *et al.* Measuring the Impact of AI in the Diagnosis of Hospitalized Patients: A Randomized Clinical Vignette Survey Study. *JAMA* **330**, 2275–2284 (2023).
16. Ghassemi, M., Oakden-Rayner, L. & Beam, A. L. The false hope of current approaches to explainable artificial intelligence in health care. *Lancet Digit. Health* **3**, e745–e750 (2021).
17. Geifman, Y. & El-Yaniv, R. Selective Classification for Deep Neural Networks. in *Neural Information Processing Systems* 4878–4887 (2017).
18. Hendrycks, D. & Gimpel, K. A Baseline for Detecting Misclassified and Out-of-Distribution Examples in Neural Networks. in *International Conference on Learning Representations* (2017).
19. Jaeger, P. F., Lüth, C. T., Klein, L. & Bungert, T. J. A Call to Reflect on Evaluation Practices for Failure Detection in Image Classification. in *The Eleventh International Conference on Learning Representations* (2023).
20. Bernhardt, M., Ribeiro, F. D. S. & Glocker, B. Failure Detection in Medical Image Classification: A Reality Check and Benchmarking Testbed. *Trans. Mach. Learn. Res.* https://openreview.net/forum?id=VBHuLfnOMf (2022).
21. Steinmetz, P., Frouin, F., Morard, V. & Buvat, I. Failure detection in biomedical image classification under realistic distribution shifts: insights from a large-scale evaluation. Preprint at https://doi.org/10.64898/2026.05.04.26350496 (2026).
22. Mamun, A. A., Khansur, M. N. O., Akash, M. A. H., Morol, M. K. & Liew, T. H. Beyond Accuracy: Quantifying Pulmonary Attribution in Anatomy-Guided Chest X-Ray Classification Under Domain Shift. Preprint at https://doi.org/10.48550/arXiv.2608.30467 (2026).
23. Badhan, M. Cethraian-Shift is a reproducible medical-imaging research project evaluating the cross-dataset reliability, calibration, label-policy sensitivity, and Grad-CAM++ stability of multi-label chest X-ray classifiers across NIH ChestX-ray14 and VinDr-CXR. Preprint at https://doi.org/10.21203/rs.3.rs-10552860/v1 (2026).
24. Zain, S., Mahamud, E., Assaduzzaman, M., Fahad, N. & Liew, T. H. Do deep chest X-ray classifiers explain where they look? A multi-architecture faithfulness benchmark with false-positive attribution audit on VinDr-CXR. *Intell.-Based Med.* **15**, 100431 (2026).
25. Ademola, J. & Bodhani, V. Explainable AI for Chest X-Ray Diagnostics: Calibration, Localization, and Human Deferral. Preprint at https://doi.org/10.5281/ZENODO.21912273 (2026).
26. Mougan, C., Broelemann, K., Kasneci, G., Tiropanis, T. & Staab, S. Explanation shift: how did the distribution shift impact the model? *Trans. Mach. Learn. Res.* https://eprints.soton.ac.uk/id/eprint/498603/ (2025).
27. Chen, J. *et al.* GAIA: Delving into Gradient-based Attribution Abnormality for Out-of-distribution Detection. in *Advances in Neural*

*Information Processing Systems 36* 79946–79958 (Neural Information Processing Systems Foundation, Inc. (NeurIPS), New Orleans, Louisiana, USA, 2023). doi:10.52202/075280-3501.

28. Huang, G., Liu, Z., Van Der Maaten, L. & Weinberger, K. Q. Densely Connected Convolutional Networks. in *2017 IEEE Conference on Computer Vision and Pattern Recognition (CVPR)* 2261–2269 (IEEE, Honolulu, HI, 2017). doi:10.1109/CVPR.2017.243.
29. Dosovitskiy, A. *et al.* An Image Is Worth 16x16 Words: Transformers for Image Recognition at Scale. in *International Conference on Learning Representations* (2021).
30. Nguyen, H. Q. *et al.* VinDr-CXR: An open dataset of chest X-rays with radiologist's annotations. *Sci. Data* **9**, 429 (2022).
31. Oakden-Rayner, L. Exploring Large-scale Public Medical Image Datasets. *Acad. Radiol.* **27**, 106–112 (2020).
32. Cambria, E., Mao, R., Chen, M., Wang, Z. & Ho, S.-B. Seven Pillars for the Future of Artificial Intelligence. *IEEE Intell. Syst.* **38**, 62–69 (2023).
33. Chefer, H., Gur, S. & Wolf, L. Transformer Interpretability Beyond Attention Visualization. in *2021 IEEE/CVF Conference on Computer Vision and Pattern Recognition (CVPR)* 782–791 (IEEE, Nashville, TN, USA, 2021). doi:10.1109/CVPR46437.2021.00084.
34. Achtibat, R. *et al.* AttnLRP: Attention-Aware Layer-Wise Relevance Propagation for Transformers. in *Proceedings of the 41st International Conference on Machine Learning* (eds Salakhutdinov, R. et al.) vol. 235 135–168 (PMLR, 2024).
35. Panboonyuen, T. Seeing Isn't Always Believing: Analysis of Grad-CAM Faithfulness and Localization Reliability in Lung Cancer CT Classification. Preprint at https://doi.org/10.48550/arXiv.2601.12826 (2026).
36. Nishankar, S., Pathirana, N., Sanjeewani, P., Perera, A. & Thuseethan, S. Does Explainability Transfer? A Controlled Benchmark of Attribution Methods on Vision Transformers and CNNs. Preprint at https://doi.org/10.48550/arXiv.2608.02396 (2026).
37. Cambria, E., Malandri, L., Mercorio, F., Mezzanzanica, M. & Nobani, N. A survey on XAI and natural language explanations. *Inf. Process. Manag.* **60**, 103111 (2023).
38. Mei, X., Mao, R., Cai, X., Yang, L. & Cambria, E. RefleXNet: Targeted Self-Reflection for Accurate Chest X-ray Reporting. *Proc. AAAI Conf. Artif. Intell.* **40**, 7954–7962 (2026).
39. Nguyen, H. Q., Pham, H. H., tuan linh, le, Dao, M. & khanh, lam. VinDr-CXR: An open dataset of chest X-rays with radiologist annotations. PhysioNet https://doi.org/10.13026/3AKN-B287 (2021).
40. Goldberger, A. L. *et al.* PhysioBank, PhysioToolkit, and PhysioNet: Components of a New Research Resource for Complex Physiologic Signals. *Circulation* **101**, e215–e220 (2000).
41. Gaggion, N. *et al.* CheXmask: a large-scale dataset of anatomical segmentation masks for multi-center chest x-ray images. *Sci. Data* **11**, 511 (2024).
42. Gaggion, N. *et al.* CheXmask Database: a large-scale dataset of anatomical segmentation masks for chest x-ray images. PhysioNet https://doi.org/10.13026/3705-ZG36 (2025).
43. Pollard, T. *et al.* PhysioNet as a global platform for biomedical research. *Nat. Health* **1**, 792–795 (2026).
44. Tejani, A. S. *et al.* Checklist for Artificial Intelligence in Medical Imaging (CLAIM): 2024 Update. *Radiol. Artif. Intell.* **6**, e240300 (2024).
45. Loshchilov, I. & Hutter, F. Decoupled Weight Decay Regularization. in *International Conference on Learning Representations* (2019).
46. Selvaraju, R. R. *et al.* Grad-CAM: Visual Explanations from Deep Networks via Gradient-Based Localization. in *2017 IEEE International Conference on Computer Vision (ICCV)* 618–626 (IEEE, Venice, 2017). doi:10.1109/ICCV.2017.74.
47. Zhou, B., Khosla, A., Lapedriza, A., Oliva, A. & Torralba, A. Learning Deep Features for Discriminative Localization. in *2016 IEEE Conference on Computer Vision and Pattern Recognition (CVPR)* 2921–2929 (IEEE, Las Vegas, NV, USA, 2016). doi:10.1109/CVPR.2016.319.
48. Ribeiro, M. T., Singh, S. & Guestrin, C. 'Why Should I Trust You?': Explaining the Predictions of Any Classifier. in *Proceedings of the 22nd ACM SIGKDD International Conference on Knowledge Discovery and Data Mining* 1135–1144 (ACM, San Francisco California USA, 2016). doi:10.1145/2939672.2939778.
49. Lundberg, S. M. & Lee, S.-I. A unified approach to interpreting model predictions. in *Proceedings of the 31st International Conference on Neural Information Processing Systems* 4768–4777 (Curran Associates Inc., Red Hook, NY, USA, 2017).
50. Erion, G., Janizek, J. D., Sturmfels, P., Lundberg, S. M. & Lee, S.-I. Improving performance of deep learning models with axiomatic attribution priors and expected gradients. *Nat. Mach. Intell.* **3**, 620–631 (2021).
51. Gaggion, N., Mansilla, L., Mosquera, C., Milone, D. H. & Ferrante, E. Improving Anatomical Plausibility in Medical Image Segmentation via Hybrid Graph Neural Networks: Applications to Chest X-Ray Analysis. *IEEE Trans. Med. Imaging* **42**, 546–556 (2023).
52. Wang, G. *et al.* Aleatoric uncertainty estimation with test-time augmentation for medical image segmentation with convolutional neural networks. *Neurocomputing* **338**, 34–45 (2019).
53. Zhang, J. *et al.* Top-Down Neural Attention by Excitation Backprop. *Int. J. Comput. Vis.* **126**, 1084–1102 (2018).
54. Otsu, N. A Threshold Selection Method from Gray-Level Histograms. *IEEE Trans. Syst. Man Cybern.* **9**, 62–66 (1979).
55. Holm, S. A Simple Sequentially Rejective Multiple Test Procedure. *Scand. J. Stat.* **6**, 65–70 (1979).
56. Chapagai, A., Gurtatta, A. S. & Cambria, E. Code and trained models for an external validation of saliency maps as error flags for chest radiograph classifiers. Zenodo https://doi.org/10.5281/ZENODO.23618485 (2026).

## Acknowledgements

We thank the NIH Clinical Center for providing ChestX-ray14, the Stanford Center for Artificial Intelligence in Medicine and Imaging and the CheXlocalize authors for the CheXpert annotations, the VinDr-CXR team and PhysioNet for VinDr-CXR, and the authors of CheXmask and HybridGNet for making their data and models available.

## Funding

The authors received no specific funding for this work.

## Author contributions

A.C. conceived and designed the study, wrote the code, performed the experiments and analyses, and wrote the manuscript. A.S.G. advised on the study design and analyses and revised the manuscript. E.C. supervised the project and revised the manuscript. All authors approved the final manuscript.

## Competing interests

The authors declare no competing interests.

## Extended Data

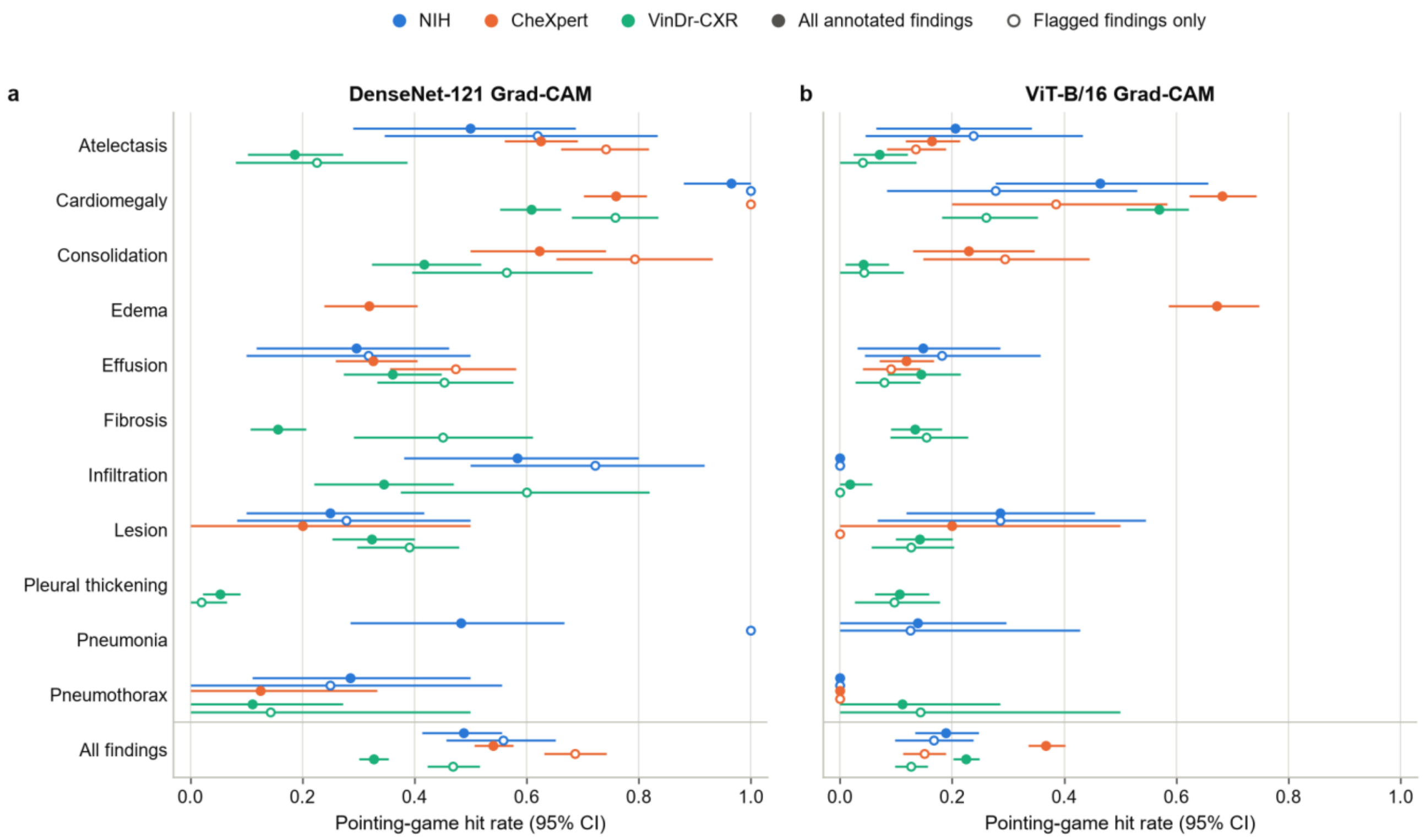

**Extended Data Fig. 1 | Localization by finding.** Pointing-game hit rate of DenseNet-121 (**a**) and ViT-B/16 (**b**) Grad-CAM by finding at NIH (blue), CheXpert (orange) and VinDr-CXR (green), over all annotated findings (filled) and over flagged findings only (open). Findings with at least five annotations at a site are shown; emphysema has no reference annotations. Bars show 95% confidence intervals (patient bootstrap; VinDr-CXR: radiographs).

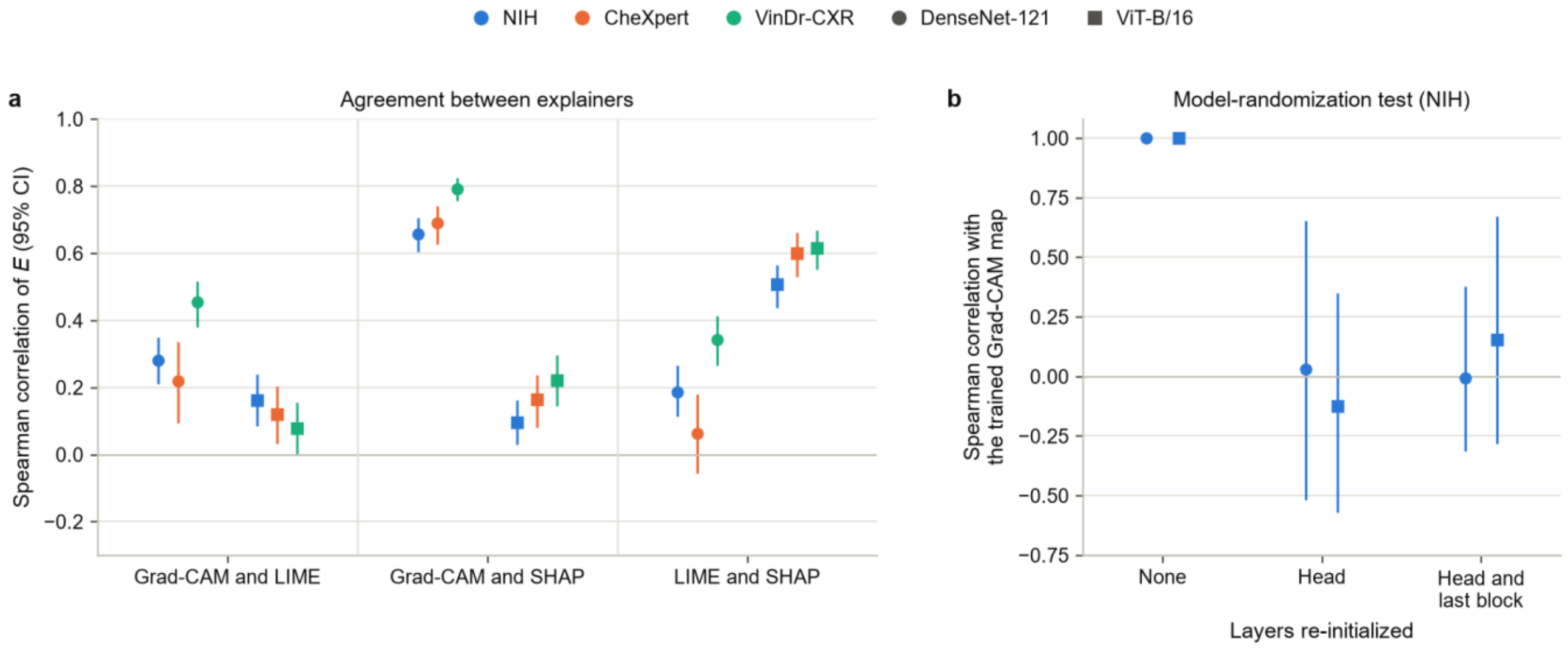

**Extended Data Fig. 2 | Agreement between explainers and model-randomization test. a**, Spearman correlation of $E$ between pairs of explainers over the same flagged findings in the cross-method subsets, by site and model, with 95% bootstrap confidence intervals. **b**, Spearman correlation between the Grad-CAM map of the trained model and the map after re-initializing the classification head, and after re-initializing the head and the last block, on 200 random NIH test radiographs (mean, with the 2.5th and 97.5th percentiles across radiographs).

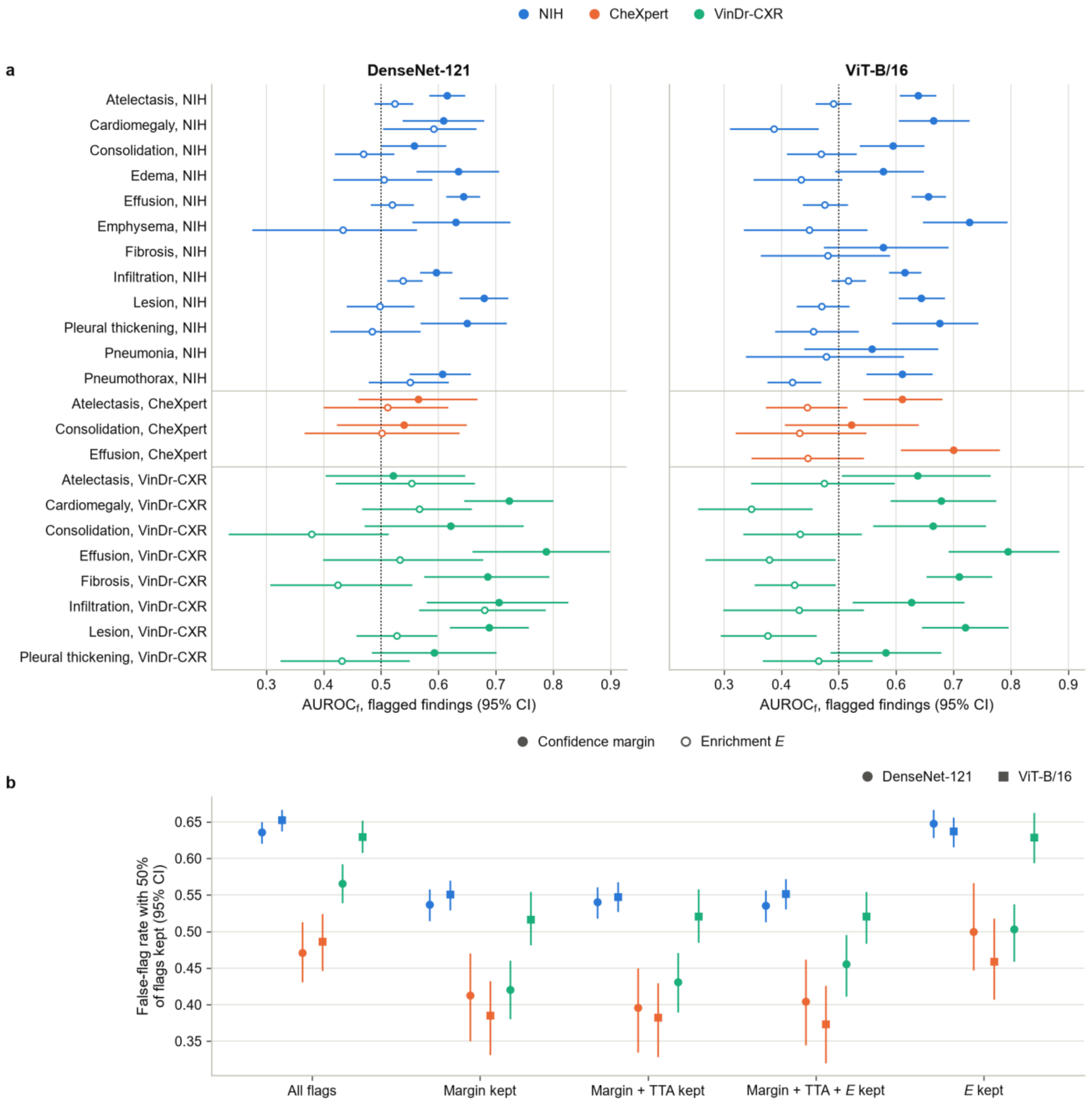

**Extended Data Fig. 3 | Failure detection by finding, and selective risk. a**, AUROC$_f$ of the absolute margin (filled) and of *E* (open) for flagged findings, by finding and site, for cells with at least 20 correct and 20 wrong flags; a row without points did not reach these counts for that model. **b**, False-flag rate among the 50% of flags with the highest score, for each score, site and model; 'All flags' means no selection. Bars show 95% confidence intervals from 1,000 bootstrap resamples of patients (VinDr-CXR: radiographs).

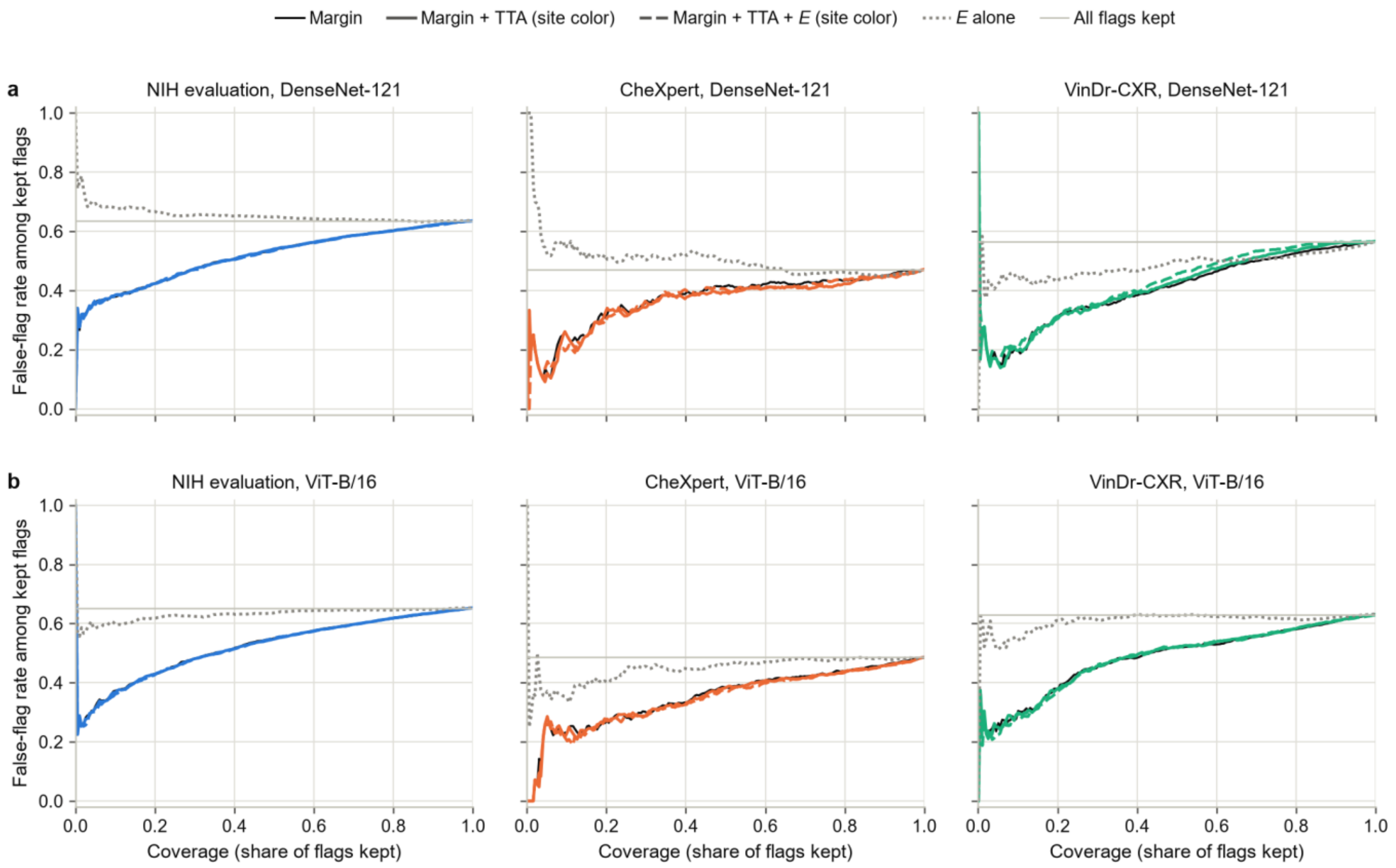

**Extended Data Fig. 4 | Risk–coverage curves for every model and site.** False-flag rate among the flags kept as flags are retained in decreasing order of each score, for DenseNet-121 (**a**) and ViT-B/16 (**b**) at NIH (evaluation half), CheXpert and VinDr-CXR. Black, margin; solid colored line, margin + TTA; dashed colored line, margin + TTA + $E$; gray dotted line, $E$ alone; light gray line, all flags kept.

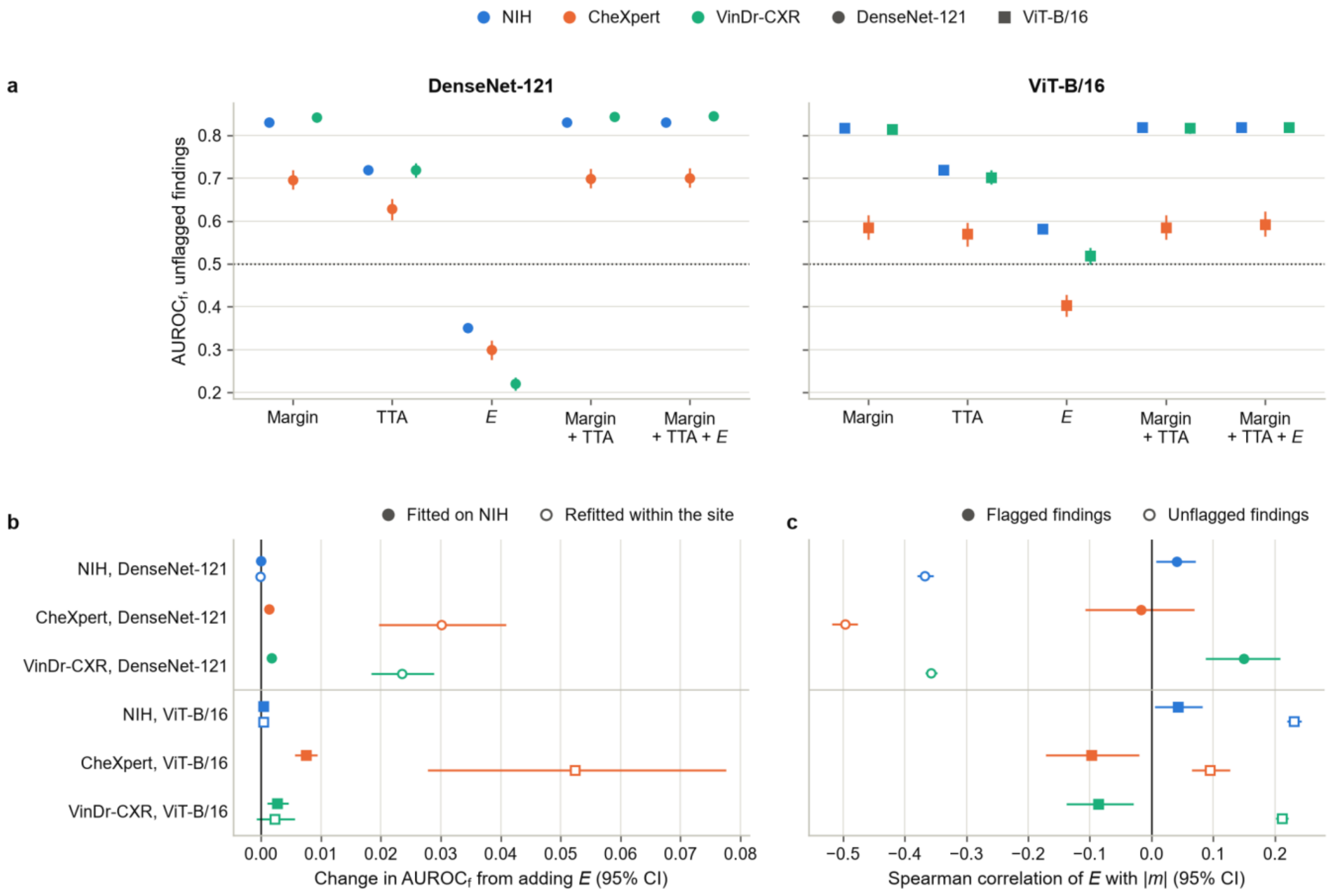

**Extended Data Fig. 5 | Missed findings and the relation between enrichment and confidence. a**, $\text{AUROC}_f$ for unflagged findings, for which a failure is a missed finding, of each score at each site. **b**, Change in $\text{AUROC}_f$ for unflagged findings from adding $E$ to the margin–TTA combination, fitted on the NIH calibration half (filled) or refitted within each site (open). **c**, Spearman correlation between $E$ and the absolute margin $|m|$ for flagged (filled) and unflagged (open) findings. Bars show 95% confidence intervals. Unflagged findings (missed findings) for DenseNet-121: 120,640 (4,092), 5,280 (595) and 34,828 (985); for ViT-B/16: 120,333 (4,144), 5,081 (500) and 34,486 (933) at NIH (evaluation half), CheXpert and VinDr-CXR.

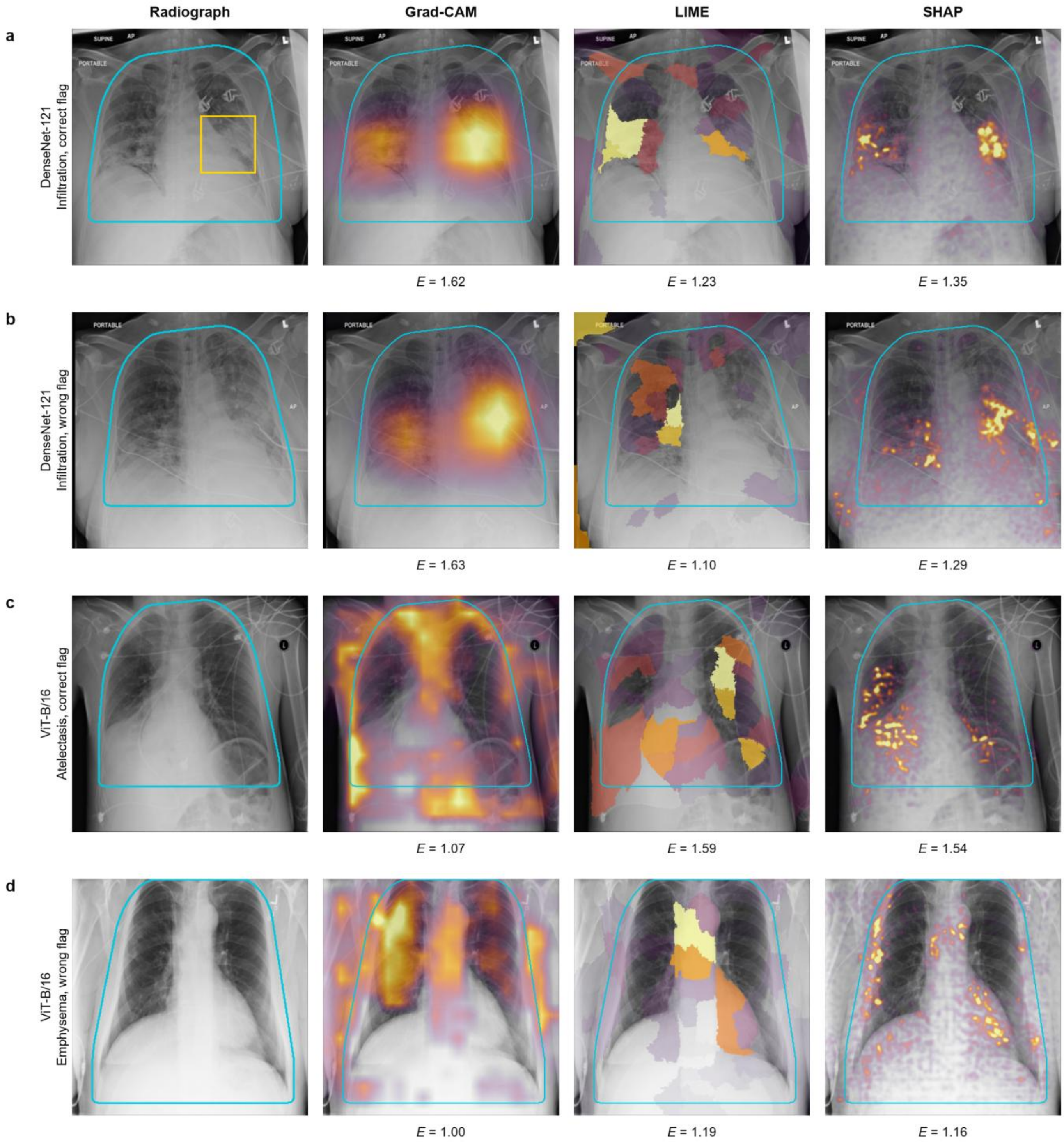

**Extended Data Fig. 6 | Grad-CAM, LIME and SHAP explanations of typical NIH flags.** For each model, the correct flag and the wrong flag in the NIH cross-method subset whose Grad-CAM enrichment is closest to the median of correct or wrong flags, respectively (**a**,**b**, DenseNet-121; **c**,**d**, ViT-B/16). Columns show the radiograph with the thoracic region $R$ (cyan) and any radiologist box (yellow), followed by positive Grad-CAM, LIME and SHAP attribution with $E$ below each map. LIME and SHAP were recomputed for the figure with the settings of the main analysis (1,000 LIME perturbations; SHAP with 50 background radiographs and 100 samples).

**Extended Data Table 1 | Internal test performance on NIH ChestX-ray14 (21,834 radiographs from 6,161 patients).**

| Metric | DenseNet-121 | ViT-B/16 | Random forest | Linear SVM |
|---|---|---|---|---|
| Macro AUROC | 0.8469 (0.8376–0.8543) | 0.8372 (0.8296–0.8439) | 0.8403 (0.8312–0.8477) | 0.8261 (0.8173–0.8340) |
| Micro AUROC | 0.8892 (0.8858–0.8925) | 0.8821 (0.8789–0.8855) | 0.8881 (0.8846–0.8913) | 0.8586 (0.8551–0.8623) |
| Macro PR-AUC | 0.2813 (0.2602–0.3028) | 0.2694 (0.2495–0.2903) | 0.2840 (0.2634–0.3058) | 0.2393 (0.2270–0.2602) |
| Macro F1 | 0.3415 (0.3186–0.3593) | 0.3213 (0.2998–0.3406) | 0.3464 (0.3247–0.3631) | 0.3115 (0.2913–0.3303) |
| Micro F1 | 0.3921 (0.3825–0.4010) | 0.3785 (0.3677–0.3882) | 0.3895 (0.3798–0.3983) | 0.3747 (0.3653–0.3838) |
| Hamming loss | 0.0721 (0.0692–0.0750) | 0.0741 (0.0713–0.0770) | 0.0770 (0.0737–0.0802) | 0.0811 (0.0780–0.0842) |
| ***AUROC by NIH class*** | | | | |
| Atelectasis | 0.8305 (0.8197–0.8405) | 0.8231 (0.8118–0.8328) | – | – |
| Cardiomegaly | 0.9193 (0.9061–0.9326) | 0.9075 (0.8914–0.9236) | – | – |
| Consolidation | 0.8114 (0.7939–0.8277) | 0.7971 (0.7803–0.8136) | – | – |
| Edema | 0.9106 (0.8969–0.9231) | 0.9091 (0.8962–0.9225) | – | – |
| Effusion | 0.8938 (0.8848–0.9025) | 0.8900 (0.8807–0.8989) | – | – |
| Emphysema | 0.9262 (0.9095–0.9402) | 0.9150 (0.8971–0.9297) | – | – |
| Fibrosis | 0.8117 (0.7879–0.8357) | 0.7985 (0.7739–0.8237) | – | – |
| Hernia | 0.8721 (0.7630–0.9512) | 0.9026 (0.8067–0.9629) | – | – |
| Infiltration | 0.7098 (0.6970–0.7219) | 0.7119 (0.6999–0.7235) | – | – |
| Mass | 0.8747 (0.8592–0.8889) | 0.8513 (0.8330–0.8684) | – | – |
| Nodule | 0.7961 (0.7782–0.8151) | 0.7434 (0.7251–0.7639) | – | – |
| Pleural thickening | 0.8259 (0.8044–0.8456) | 0.8113 (0.7906–0.8313) | – | – |
| Pneumonia | 0.7851 (0.7567–0.8137) | 0.7796 (0.7490–0.8089) | – | – |
| Pneumothorax | 0.8891 (0.8724–0.9041) | 0.8805 (0.8625–0.8972) | – | – |

Values are estimates with 95% confidence intervals from 1,000 bootstrap resamples of patients. F1 scores and Hamming loss are at the macro-F1 thresholds tuned on validation data. The random forest and the linear support vector machine (SVM) use the pooled 1,024-dimensional DenseNet-121 features. AUROC, area under the receiver operating characteristic curve; PR-AUC, area under the precision–recall curve.

**Extended Data Table 2 | Shift in Grad-CAM enrichment of flagged findings relative to NIH.**

| Finding | Flags, NIH / external | Median *E*, NIH | Median *E*, external | Difference (95% CI) | *P* | Holm *P* |
|---|---|---|---|---|---|---|
| ***DenseNet-121, CheXpert versus NIH*** | | | | | | |
| All findings (pooled) | 19,537 / 480 | 1.601 | 1.592 | −0.010 (−0.042 to 0.028) | 0.577 | – |
| Atelectasis | 3,079 / 153 | 1.700 | 1.590 | −0.110 (−0.160 to −0.061) | $1.7 \times 10^{-6}$ | $8.6 \times 10^{-6}$ |
| Cardiomegaly | 664 / 34 | 1.935 | 1.705 | −0.230 (−0.321 to −0.100) | $8.7 \times 10^{-6}$ | $3.5 \times 10^{-5}$ |
| Consolidation | 2,058 / 106 | 1.813 | 1.703 | −0.110 (−0.179 to −0.024) | 0.002 | 0.003 |
| Effusion | 2,720 / 93 | 1.618 | 1.467 | −0.151 (−0.186 to −0.119) | $3.5 \times 10^{-10}$ | $2.1 \times 10^{-9}$ |
| Lesion | 2,232 / 61 | 1.620 | 1.638 | 0.018 (−0.086 to 0.062) | 0.439 | 0.439 |
| Pneumothorax | 1,393 / 26 | 1.512 | 1.318 | −0.193 (−0.325 to −0.133) | $3.9 \times 10^{-5}$ | $1.2 \times 10^{-4}$ |
| ***DenseNet-121, VinDr-CXR versus NIH*** | | | | | | |
| All findings (pooled) | 19,537 / 1,160 | 1.601 | 1.591 | −0.011 (−0.049 to 0.032) | 0.309 | – |
| Atelectasis | 3,079 / 118 | 1.700 | 1.643 | −0.057 (−0.103 to 0.046) | 0.206 | 1.000 |
| Cardiomegaly | 664 / 179 | 1.935 | 1.938 | 0.004 (−0.082 to 0.071) | 0.522 | 1.000 |
| Consolidation | 2,058 / 67 | 1.813 | 1.712 | −0.101 (−0.188 to 0.000) | 0.003 | 0.023 |
| Effusion | 2,720 / 84 | 1.618 | 1.527 | −0.091 (−0.132 to 0.004) | 0.078 | 0.471 |
| Fibrosis | 194 / 82 | 1.553 | 1.494 | −0.059 (−0.121 to 0.009) | 0.282 | 1.000 |
| Infiltration | 4,263 / 195 | 1.441 | 1.334 | −0.107 (−0.171 to −0.046) | $5.4 \times 10^{-8}$ | $5.4 \times 10^{-7}$ |
| Lesion | 2,232 / 258 | 1.620 | 1.645 | 0.025 (−0.033 to 0.081) | 0.878 | 1.000 |
| Pleural thickening | 1,040 / 107 | 1.562 | 1.470 | −0.092 (−0.146 to 0.016) | 0.004 | 0.029 |
| Pneumonia | 396 / 26 | 1.735 | 1.767 | 0.032 (−0.144 to 0.120) | 0.830 | 1.000 |
| Pneumothorax | 1,393 / 27 | 1.512 | 1.259 | −0.253 (−0.374 to −0.130) | $6.2 \times 10^{-5}$ | $5.6 \times 10^{-4}$ |
| ***ViT-B/16, CheXpert versus NIH*** | | | | | | |
| All findings (pooled) | 19,860 / 679 | 0.993 | 1.196 | 0.203 (0.172 to 0.234) | $<10^{-16}$ | – |
| Atelectasis | 3,136 / 263 | 1.175 | 1.140 | −0.035 (−0.076 to −0.006) | 0.002 | 0.008 |
| Cardiomegaly | 884 / 28 | 1.370 | 1.331 | −0.039 (−0.121 to 0.064) | 0.334 | 0.668 |
| Consolidation | 1,443 / 111 | 1.039 | 1.090 | 0.051 (0.000 to 0.096) | 0.086 | 0.346 |
| Effusion | 2,973 / 154 | 1.525 | 1.529 | 0.004 (−0.063 to 0.049) | 0.991 | 0.991 |
| Lesion | 1,817 / 52 | 0.924 | 1.181 | 0.257 (0.149 to 0.310) | $1.8 \times 10^{-10}$ | $1.1 \times 10^{-9}$ |
| Pneumothorax | 1,204 / 54 | 1.156 | 1.233 | 0.077 (−0.024 to 0.198) | 0.207 | 0.621 |
| ***ViT-B/16, VinDr-CXR versus NIH*** | | | | | | |
| All findings (pooled) | 19,860 / 1,502 | 0.993 | 1.028 | 0.035 (0.007 to 0.076) | 0.158 | – |
| Atelectasis | 3,136 / 129 | 1.175 | 1.126 | −0.048 (−0.094 to 0.016) | 0.209 | 0.626 |
| Cardiomegaly | 884 / 136 | 1.370 | 1.403 | 0.032 (−0.085 to 0.104) | 0.796 | 1.000 |
| Consolidation | 1,443 / 123 | 1.039 | 0.851 | −0.188 (−0.231 to −0.125) | $1.7 \times 10^{-11}$ | $1.5 \times 10^{-10}$ |
| Effusion | 2,973 / 103 | 1.525 | 1.363 | −0.162 (−0.238 to −0.035) | 0.002 | 0.012 |
| Fibrosis | 678 / 366 | 1.032 | 1.131 | 0.099 (0.033 to 0.149) | $3.0 \times 10^{-6}$ | $2.1 \times 10^{-5}$ |
| Infiltration | 4,838 / 216 | 0.576 | 0.426 | −0.150 (−0.177 to −0.083) | $2.2 \times 10^{-15}$ | $2.2 \times 10^{-14}$ |
| Lesion | 1,817 / 197 | 0.924 | 1.015 | 0.091 (0.059 to 0.142) | $4.4 \times 10^{-6}$ | $2.6 \times 10^{-5}$ |
| Pleural thickening | 900 / 140 | 1.109 | 1.076 | −0.032 (−0.111 to 0.042) | 0.106 | 0.422 |
| Pneumonia | 775 / 54 | 0.607 | 0.459 | −0.148 (−0.222 to −0.077) | $2.1 \times 10^{-8}$ | $1.7 \times 10^{-7}$ |
| Pneumothorax | 1,204 / 24 | 1.156 | 1.317 | 0.162 (−0.210 to 0.309) | 0.508 | 1.000 |

Differences of medians (external minus NIH) with 95% bootstrap confidence intervals (patients resampled within each site; VinDr-CXR: radiographs) and two-sided Mann–Whitney *P* values; Holm adjustment within each model and external site across the findings tested. Findings with at least 20 flags at both sites are shown.

**Extended Data Table 3 | Failure detection for flagged and unflagged findings.**

| Site | Decisions (failures) | Margin | TTA | *E* | M + T | M + T + *E* | ΔAUROC$_f$ (95% CI) | ΔAURC (95% CI) |
|---|---|---|---|---|---|---|---|---|
| ***DenseNet-121, flagged findings*** | | | | | | | | |
| NIH (evaluation half) | 9,296 (5,908) | 0.646 | 0.549 | 0.484 | 0.646 | 0.648 | +0.0022 (−0.0011 to 0.0052) | −0.0009 (−0.0025 to 0.0009) |
| CheXpert | 480 (226) | 0.607 | 0.616 | 0.487 | 0.625 | 0.616 | −0.0087 (−0.0205 to 0.0044) | +0.0022 (−0.0052 to 0.0086) |
| VinDr-CXR | 1,160 (656) | 0.685 | 0.489 | 0.618 | 0.669 | 0.646 | −0.0233 (−0.0315 to −0.0155) | +0.0128 (0.0088 to 0.0168) |
| ***DenseNet-121, unflagged findings*** | | | | | | | | |
| NIH (evaluation half) | 120,640 (4,092) | 0.830 | 0.720 | 0.351 | 0.831 | 0.831 | 0.0000 (−0.0001 to 0.0001) | 0.0000 (0.0000 to 0.0000) |
| CheXpert | 5,280 (595) | 0.697 | 0.629 | 0.300 | 0.699 | 0.700 | +0.0014 (0.0011 to 0.0018) | −0.0003 (−0.0004 to −0.0002) |
| VinDr-CXR | 34,828 (985) | 0.842 | 0.719 | 0.221 | 0.844 | 0.845 | +0.0018 (0.0015 to 0.0020) | −0.0001 (−0.0001 to −0.0001) |
| ***ViT-B/16, flagged findings*** | | | | | | | | |
| NIH (evaluation half) | 9,603 (6,267) | 0.660 | 0.579 | 0.524 | 0.661 | 0.661 | +0.0002 (−0.0023 to 0.0024) | −0.0012 (−0.0023 to 0.0000) |
| CheXpert | 679 (330) | 0.656 | 0.612 | 0.536 | 0.664 | 0.668 | +0.0048 (−0.0004 to 0.0104) | −0.0038 (−0.0070 to −0.0008) |
| VinDr-CXR | 1,502 (946) | 0.681 | 0.550 | 0.522 | 0.676 | 0.678 | +0.0017 (−0.0037 to 0.0068) | −0.0031 (−0.0057 to −0.0005) |
| ***ViT-B/16, unflagged findings*** | | | | | | | | |
| NIH (evaluation half) | 120,333 (4,144) | 0.817 | 0.719 | 0.582 | 0.818 | 0.819 | +0.0004 (−0.0002 to 0.0011) | 0.0000 (−0.0001 to 0.0000) |
| CheXpert | 5,081 (500) | 0.585 | 0.570 | 0.403 | 0.585 | 0.593 | +0.0076 (0.0057 to 0.0094) | −0.0013 (−0.0018 to −0.0008) |
| VinDr-CXR | 34,486 (933) | 0.814 | 0.702 | 0.519 | 0.817 | 0.819 | +0.0027 (0.0011 to 0.0046) | −0.0001 (−0.0002 to 0.0000) |

AUROC$_f$ of each score (0.5, uninformative). The combinations were fitted on the NIH calibration half and evaluated on the NIH evaluation half and the external sites. ΔAUROC$_f$ and ΔAURC compare the combination with and without *E* on the same bootstrap resamples (negative ΔAURC is an improvement). TTA, test-time augmentation agreement; M + T, margin + TTA; M + T + *E*, margin + TTA + enrichment.

**Extended Data Table 4 | Anatomical reference: mask quality, agreement with CheXmask and coverage of radiologist annotations.**

| Site | Masks used | Lung Dice vs CheXmask | Heart Dice vs CheXmask | Annotation area inside *R* |
|---|---|---|---|---|
| NIH | CheXmask (HybridGNet); 21,564 of 21,834 (98.8%) pass | 0.997 (0.996); *n* = 300 | 0.998 (0.997); *n* = 300 | 96.3% (100%; 77.3%); 188 boxes |
| CheXpert | Our HybridGNet run; 720 of 720 pass | 0.995 (0.988); *n* = 200 | 0.995 (0.991); *n* = 200 | 99.0% (100%; 94.0%); 827 segmentations |
| VinDr-CXR | Our HybridGNet run; 2,999 of 3,000 pass | 0.997 (0.996); *n* = 3,000 | 0.997 (0.995); *n* = 3,000 | 98.0% (100%; 85.9%); 1,660 boxes |

NIH masks are the released CheXmask masks with an estimated reverse-classification-accuracy Dice of at least 0.7; CheXpert and VinDr-CXR masks were produced with the released HybridGNet weights and passed the plausibility check described in Methods. Agreement compares our HybridGNet masks with the released CheXmask masks on the radiographs that CheXmask covers (mean Dice, with the 5th percentile in parentheses). Coverage is the share of each annotation's area inside the thoracic region *R* (mean, with the median and the 5th percentile in parentheses).